\documentclass{article}
\usepackage[utf8]{inputenc}
\usepackage{graphicx}
\usepackage{mathrsfs} 
\usepackage{subfigure}
\usepackage{pgfplots}
\usepgfplotslibrary{groupplots}
\pgfplotsset{compat=1.12}
\usepackage{tikz}
\usepackage{tikzscale}
\usepackage{algorithm}
\usepackage[noend]{algpseudocode}
\usepackage{amsmath}
\usepackage{stfloats}
\usepackage[a4paper,margin=1in,footskip=0.25in]{geometry}

\title{GRIHA: Synthesizing 2-Dimensional Building Layouts from Images Captured using a Smart Phone}
\author{Shreya Goyal, Naimul Khan, Chiranjoy Chattopadhyay, and Gaurav Bhatnagar}
\date{March 2021}

\begin{document}

\maketitle

\begin{abstract}
Reconstructing an indoor scene and generating a layout/floor plan in 3D or 2D is a widely known problem. Quite a few algorithms have been proposed in the literature recently. However, most existing methods either use RGB-D images, thus requiring a depth camera, or depending on panoramic photos, assuming that there is little to no occlusion in the rooms. In this work, we proposed GRIHA (Generating Room Interior of a House using  ARCore), a framework for generating a layout using an RGB image captured using a simple mobile phone camera. We take advantage of Simultaneous Localization and Mapping (SLAM) to assess the 3D transformations required for layout generation. SLAM technology is built-in in recent mobile libraries such as ARCore by Google.  Hence, the proposed method is fast and efficient. It gives the user freedom to generate layout by merely taking a few conventional photos, rather than relying on specialized depth hardware or occlusion-free panoramic images. We have compared GRIHA with other existing methods and obtained superior results. Moreover, the system is tested on multiple hardware platforms to test the dependency and efficiency.
\end{abstract}

\section{Introduction}

\label{sec:intro}
Generating Building Information Model (BIM) in 2D/3D from indoor scenes has applications including real estate websites, indoor navigation, augmented/virtual reality, and many more. BIM should include the global layout of the entire floor plan of the space, number of rooms, and arrangements. The most natural way to create the floor plan is by manually measuring each room and integrating everything in Computer-Aided Design (CAD) software to have a global layout. However, such manual measurement is a tedious task and requires a significant amount of work hours. Hence, there is a need to develop a graphics and visualization system that takes a few images or videos of the entire indoor scene and automatically generates a floor plan. The pictures or videos may or may not have depth information since a typical user might not have a depth camera or specialized hardware to measure depth in the scene.

\begin{figure}[t]
    \centering
    \includegraphics[width=\linewidth]{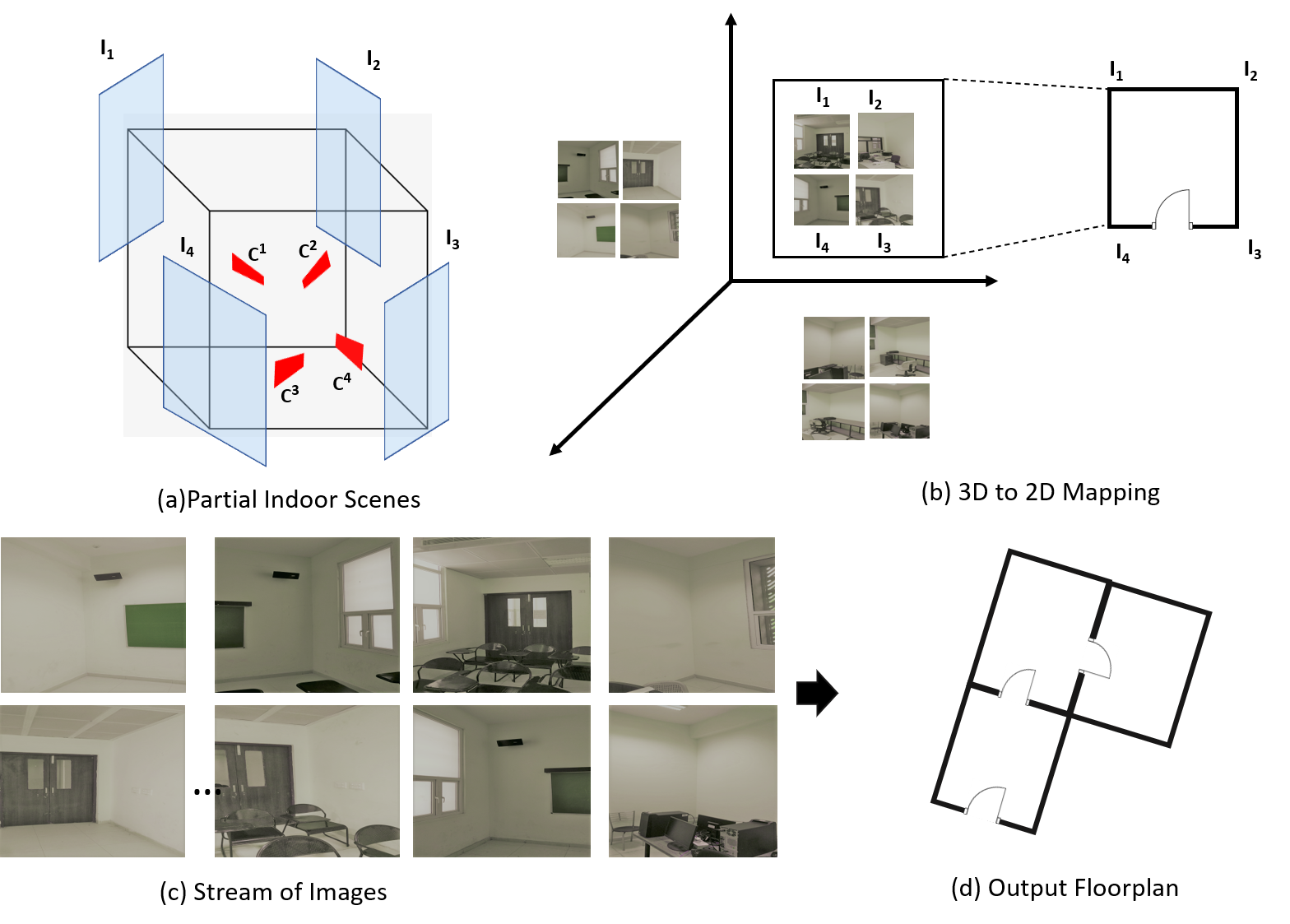}
    \caption{A schematic illustration of the entire input and output scenario while using GRIHA.}
    \label{fig:prob}
\end{figure}

However, estimating an accurate layout from RGB images is a challenging task. It requires precise depth information along with the RGB images and an exact point cloud with real-world scaling. The current state of scholarship (as discussed in Sec. \ref{sec:litsurv}), as well as commercial applications for layout estimation, uses RGB-D images or panorama images to solve this problem. Despite good accuracy, these methods require special hardware (depth camera) or a particular photo capture mode (panorama with little to no occlusion). Such requirements restrict widespread adoption.  In this work, we propose a framework named GRIHA (Generating Room Interior of a House using  ARCore). This a framework to estimate the 2D floor plan of an indoor scene using RGB images captured through a mobile camera.  Figure \ref{fig:prob} depicts the potential partial input RGB images of indoor locations and their corresponding floor plan of the entire indoor scene. In our proposed method, the user needs to take pictures of all the indoor scene rooms using our developed application. The depth map of the locations is extracted from these pictures adapting the method in \cite{alhashim2018high} followed by edge map extraction using \cite{zou2018layoutnet}. The RGB images and depth maps are used for 3D reconstruction. The point clouds and edge maps are then mapped in 2D for the final layout using a novel regularization technique. We take advantage of Google ARCore library for pose estimation, thus saving computational time. 

In this paper, our major contributions are: (1) 2D floor plan estimation from RGB images without using panorama or $360^{\circ}$ spherical images, (2) Development of an ARcore based Android mobile application for data collection and application of ARcore pose data for 3D reconstruction of the scene, (3) Near accurate layout estimation with a fewer number of images captured through the camera, compared to the existing techniques, and (4) Cross-platform testing of the proposed framework on multiple devices. Currently, we are assuming only Manhattan and weak Manhattan scenes. The proposed method compares well in terms of dimensions measured by the existing commercial applications on the Android platform to generate the layout and measure the dimensions.  GRIHA works well in case of occlusion in the indoor scene where existing methods may fail. 

The remaining of this manuscript is organized as: Section \ref{sec:litsurv} highlights the main differences between GRIHA and the state-of-the-art. Section \ref{sec:prob} defines the problem statement and Sec. \ref{sec:data_collection} gives the overview of the data collection step for the proposed system.  Section \ref{sec:system} describes the proposed system and its constituent steps. Section \ref{sec:exp} describes the various qualitative and quantitative experiments performed for the proposed algorithm and sub-modules, and Sec. \ref{sec:conclusion} concludes with an outlook for the future work.

\section{Related Work}
\label{sec:litsurv}
Layout estimation from RGB-D and panorama images has been a widely explored. The authors in \cite{liu2015rent3d,zhang2013estimating} have reconstructed the indoor scene in 3D using monocular images and estimated the layout using vanishing points and depth features. In \cite{zhang2019edge}, authors have calculated layouts on RGB images by an encoder-decoder framework to jointly learn the edge maps and semantic labels. In  \cite{bao2014understanding} room layout is generated from pictures taken from multiple views and reconstructed using SfM and region classification. In another work,\cite{dasgupta2016delay} authors estimated layout in a cluttered indoor scene by identifying label for a pixel from RGB images, using deep FCNN, and refined using geometrical techniques. 

Since monocular images cannot capture the entire scene, layout from panorama images to increase, view is an alternative. In \cite{zou2018layoutnet}, an encoder-decoder network predicts boundary and corner maps and optimized over constrained geometrical properties of the room. Authors in \cite{sun2019horizonnet} estimated room layout by regressing over boundaries and classifying the corners for each column representation of an image. Furthermore, \cite{fernandez2018layouts} generated a 3D structure for 360-degree panorama image by reasoning between geometry and edge maps returned by a deep neural network. The work presented in \cite{hsiao2019flat2layout} proposes an approach to generate indoor scene layout by learning the encoded layout representation in row vectors. 
In \cite{dong2015imoon}, the authors have generate a navigation mesh with captured point clouds using SfM. In \cite{xu2017pano2cad,zhang2014panocontext}, layout is obtained from a panorama image by estimating object locations and pose in the room. The indoor layout has also been generated from monocular video sequences in \cite{furlan2013free}, where SLAM/SfM techniques are used for 3D reconstruction. In another work \cite{angladon2018room}, the author has generated a 2D layout of the room using depth data and 3D reconstruction using SLAM.  In the context of floor plan generation, \cite{cabral2014piecewise} has reconstructed the layout from the input panorama images using SfM and generated a 2D floor plan by posing it as the shortest path problem. In addition, \cite{lin2018floorplan} predicts the global room layout and transformations using partial reconstructions of indoor scenes using RGB-D images without using feature matching between partial scans. In another work, \cite{phalak2020scan2plan} proposed a floor plan estimation method from 3D indoor scans by offering a 2 stage method, where the first-stage clusters the wall and room instances the second stage predicts the perimeter of the rooms. 

In \cite{murali2017indoor}, a BIM of the interior of the house was generated from 3D scans. In \cite{chen2015crowd}, authors proposed CrowdMap, which used sensor-rich video data for reconstructing floor plans from indoor scenes. Authors in \cite{turner2014floor} have generated floor plans by triangulating the space between wall samples and partitioning space into the interior and exterior spaces. In \cite{chelani2018towards}, 2D floor plans are generated by estimating camera motion trajectories from the depth and RGB sequences and by registering 3D data into a simplified rectilinear representation. In \cite{okorn2010toward}, authors have proposed a system to create 2D floor plans for building interiors which takes 3D point clouds from 3D scanners by utilizing height data for detecting wall and ceiling points and projecting the rest of the points in a 2D plane. In the work proposed by \cite{murez2020atlas}, authors have proposed an end to end framework for 3D reconstruction of a scene using RGB images by directly regressing to 3D. The existing layout estimation methods in the literature largely depends upon the depth-sensing hardware and occlusion free panorama. Hence, we propose a layout estimation system, GRIHA, which uses multiple 2D images of an indoor scene, taken from a traditional mobile phone's camera, which suppresses the requirement of depth-sensing hardware and easier to use in the occluded indoor environment.

\begin{figure*}[!b]
    \centering
    \includegraphics[width=\linewidth]{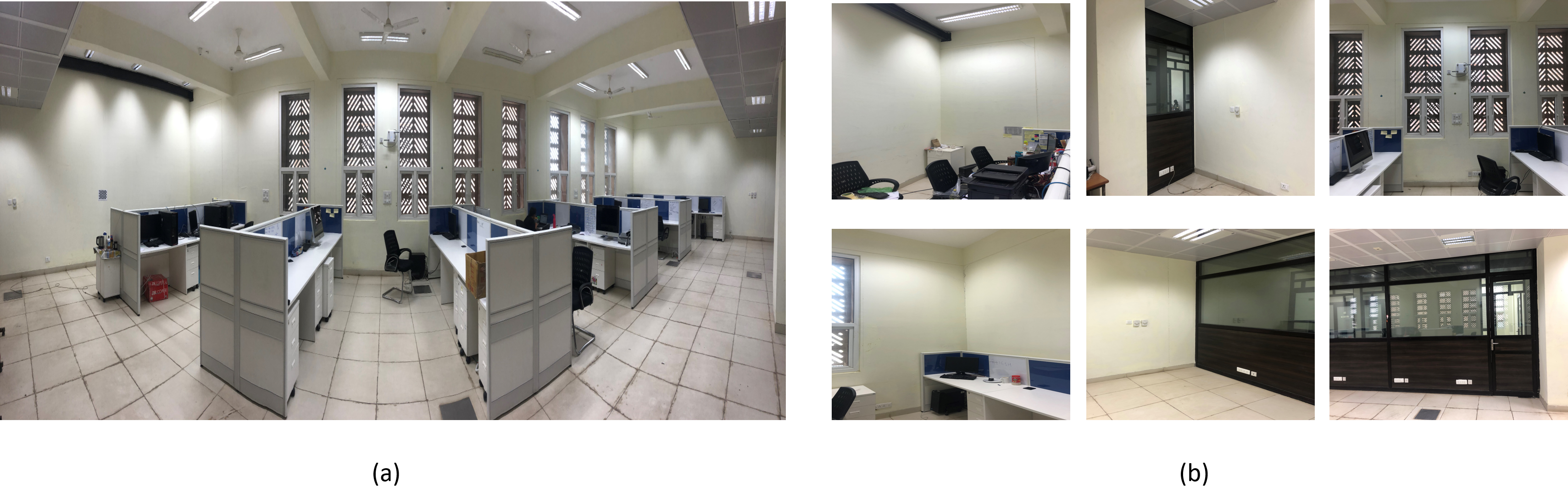}
    \caption{Panoramic image and multiple 2D images for the same scene.}
    \label{fig:panoramavs2D}
\end{figure*}
\section{Problem Statement}
\label{sec:prob}
Figure \ref{fig:prob} shows the input and output of the proposed system GRIHA. Figure \ref{fig:prob}(a) shows the partial images taken from every corner of the room by mobile phone's camera. Here, $C^1$, $C^2$, $C^3$, $C^4$ are the four locations from which a user captures four room images using a mobile camera. These captured images are marked as $I_1$, $I_2$, $I_3$, $I_4$, where image $I_k$ is captured from camera pose $C^k$. These are the images that act as inputs to the GRIHA framework. Figure. \ref{fig:prob} (b) depicts every corner image’s mapping to a partial 2D layout. To construct the 2D floor plan, we need to map the images to the corresponding corner points of a room in a Manhattan world and establish a sequence to get the room’s complete layout. Figure. \ref{fig:prob} (c), and (d) show the input stream of indoor scene images for a complete floor and its respective output floor plan. To summarize, the task at hand is \textit{Given a sequence of images of all the rooms (for every room, images of four corners and the door) in a single floor of a building, generate the 2-D plan of the entire floor}.

\begin{figure*}[t]
    \centering
    \includegraphics[width=\linewidth]{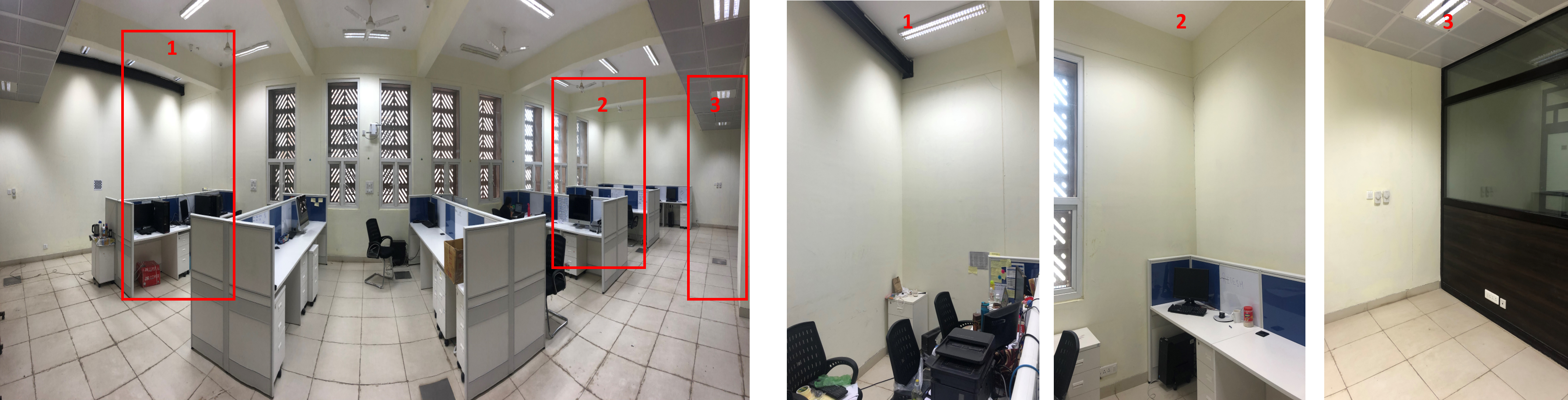}
    \caption{Highlighted occluded portions of the scene and corresponding 2D images.}
    \label{fig:panoramavs2D1}
\end{figure*}

\section{Image Acquisition Platform}
\label{sec:data_collection}
Most of the methods for 3D reconstruction of a room and floor plan estimation requires an ample amount of specific hardware such as depth camera, Kinect camera, LiDAR. Although some methods exist for layout generation from monocular images \cite{zou2018layoutnet}, they rely on occlusion-free panoramic photos, which are almost impossible to take in common spaces such as an office or home. Instead, our proposed method relies on multiple regular 2D photos, which are easier to capture in an occlusion free manner. Figure \ref{fig:panoramavs2D} (a) shows the panoramic image of a large office space and Fig. \ref{fig:panoramavs2D}(b) shows multiple 2D images for the same scene. 

\begin{figure}[!b]
    \centering
    \includegraphics[width=\linewidth]{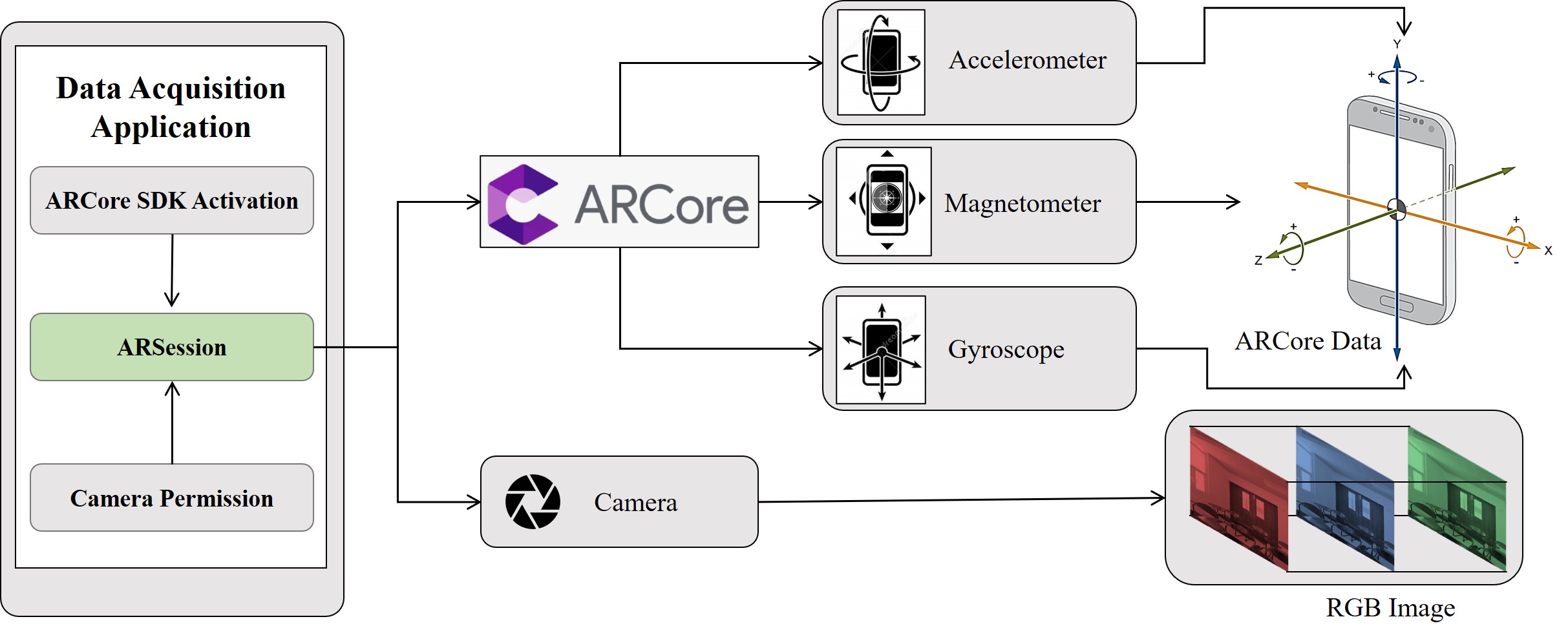}
    \caption{A schematic representation of the image acquisition framework developed for GRIHA.}
    \label{fig:data}
\end{figure}

It can be seen that the panoramic image of a vast space may have several occlusions, which make their 3D reconstruction difficult and erroneous. Most of their corners and other important edges may get occluded in the panorama because of the furniture and other decors. At the same time, they are easier to capture in their multiple 2D images. It’s not possible to cover the complete scene (backside of the camera) in a panoramic image without losing important information. While capturing multiple 2D images makes it easier to capture all the corners, edges, and other portions of the scene, without getting occluded by the furniture and without losing any information. Figure \ref{fig:panoramavs2D1} shows the highlighted portions of the scene in the panoramic image and their corresponding 2D images, which have more information about the scene than their panoramic version. Hence, generating the scene’s layout from their separate 2D images gives more information about the scene than generating it from its panoramic image.

\begin{figure*}[hbt!]
    \centering
    \includegraphics[width=\linewidth]{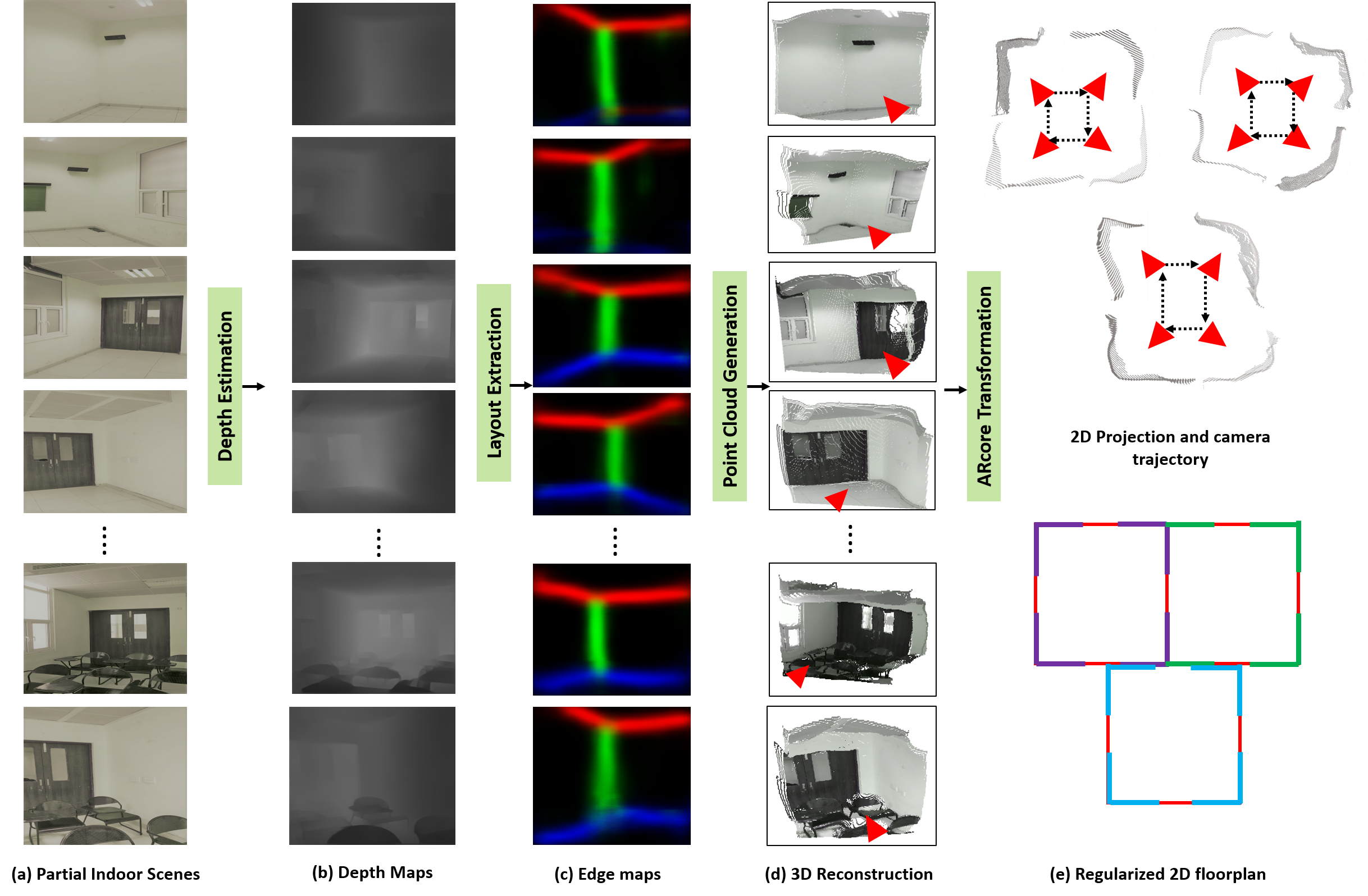}
    \caption{Pipeline of the proposed framework.}
    \label{fig:framework}
\end{figure*}

We used smartphones to capture images and track camera motion for our experiments. The proposed method attempts to address this issue by taking advantage of recent mobile Augmented Reality libraries for pose estimation. Figure \ref{fig:data} shows an illustration of how GRIHA collects data using ARCore, which readily available on almost all Android devices. ARcore \cite{arcore} is a library by Google, which uses the phone's IMU sensor's data along with image feature points for tracking the pose of the camera utilizing a proprietary Simultaneous Localization and Mapping (SLAM) algorithm. It performs pose estimation in real-time, challenging to achieve with a third-party SLAM algorithm. To collect the data, an android application was developed in Unity3D environment. The camera’s real-world position and orientation were extracted. The proposed method takes partial scene images for each room in the house for floor plan reconstruction.

\section{Proposed System}
\label{sec:system}
\subsection{Overview of GRIHA}
\label{sec:overview}
 Figure \ref{fig:framework} depicts a step-by-step visual illustration of the working principle of GRIHA. As shown in Fig. \ref{fig:framework}(a), a stream of RGB images are taken for each partial scene in the entire house. Depth estimation is done individually for each RGB image (see Fig. \ref{fig:framework}(b)) and edge/boundary map is estimated by adapting the network proposed in \cite{zou2018layoutnet} for perspective images (see Fig. \ref{fig:framework}(c)). Mapping the depth maps with RGB images and edge maps of layouts, the point clouds are generated and layout is extracted from the point clouds (see Fig. \ref{fig:framework}(d)). When the pictures are being taken, we utilized ARCore's built-in SLAM algorithm to extract the 3D pose of the camera and saved it alongside the images. The camera poses are utilized to rebuild the camera trajectory, which is utilized for transforming each partial scene point cloud after projecting them into a 2D plane. The point cloud of the layout is extracted from the 3D reconstructed scene using edge maps generated in the previous steps. All the partial point clouds are projected in 2D and translated using ARcore transformations. Each partial point cloud is then regularized locally and globally to generate the 2D floor plan for the house (see Fig. \ref{fig:framework}(e)). The red markers indicate the trajectory of the camera while capturing the images for each room in the house.

\subsection{Depth Estimation} 
\label{sec:depth}
\begin{figure}[t]
    \centering
    \includegraphics[width=\linewidth]{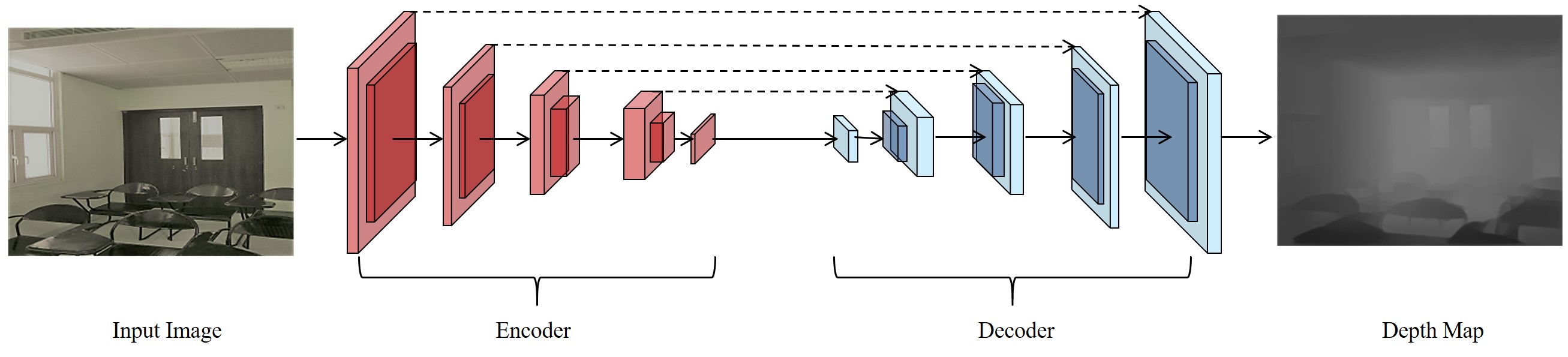}
    \caption{An illustration of how depth is estimated by adapting the encoder-decoder model proposed in \cite{alhashim2018high}.}
    \label{fig:depth}
\end{figure}

\begin{figure}[!b]
    \centering
    \includegraphics[width=\linewidth]{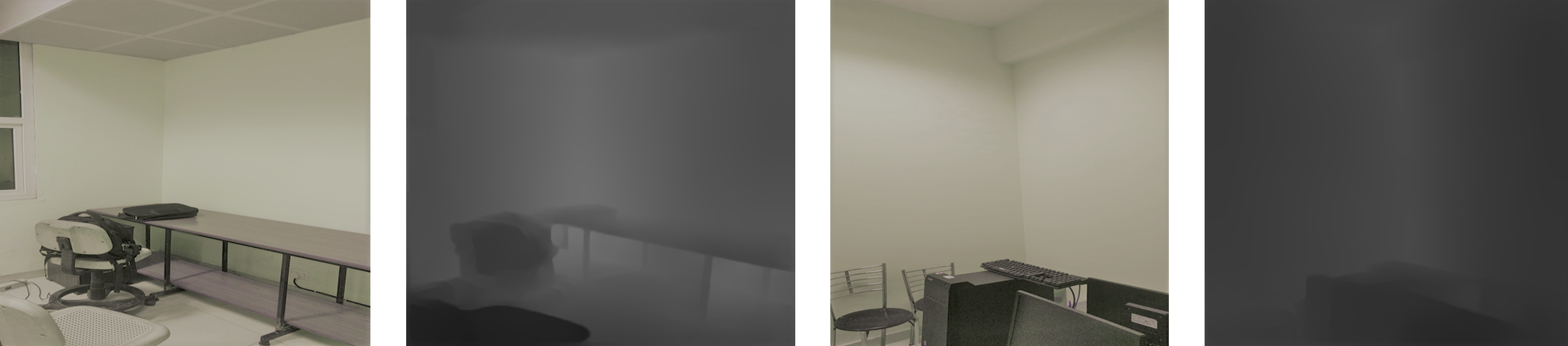}
    \caption{Results of depth generation on our dataset.}
    \label{fig:depth_res}
\end{figure}

Depth perception is essential for estimating the correct dimension of the targeted floor plan. Figure \ref{fig:depth} depicts an illustration of how the depth map is computed from a given image. Traditionally, a device with a built-in depth camera such as Google Tango or Microsoft Kinect is used for capturing point clouds directly from the scene. However, in GRIHA, the input taken is an RGB image taken from a traditional mobile device’s camera. Hence, depth estimation is a mandatory step in this context. For depth perception from RGB images, multiple methods exploit feature matching techniques in multiple images of the same scene and reconstruct a 3D model for it. However, such schemes require a trained user to capture the data to ensure correspondence across images. Hence, a feasible solution for estimating depth from a single image is to use a pre-trained machine learning model. Depth for RGB images can be learned in a supervised manner from ground truth depth-maps and a trained neural network can be used for estimating depth for new images. We have adapted the method proposed in \cite{alhashim2018high}, where an encoder-decoder architecture is used after extracting image features with DenseNet-169, resulting in high-resolution depth-maps. The encoder used in this method is a pre-trained truncated DenseNet-169. The decoder consists of basic blocks of convolutional layers, concatenated with successive $2\times$ bilinear upsampling blocks, and two $3 \times 3$ convolutional layers, where the output filter is half the size of the input. Figure. \ref{fig:depth_res} shows the results for the depth estimation model on our dataset. The depth estimation model’s performance analysis is present in Sec. \ref{sec:exp}. 

\begin{figure}[t]
    \centering
    \includegraphics[width=\linewidth]{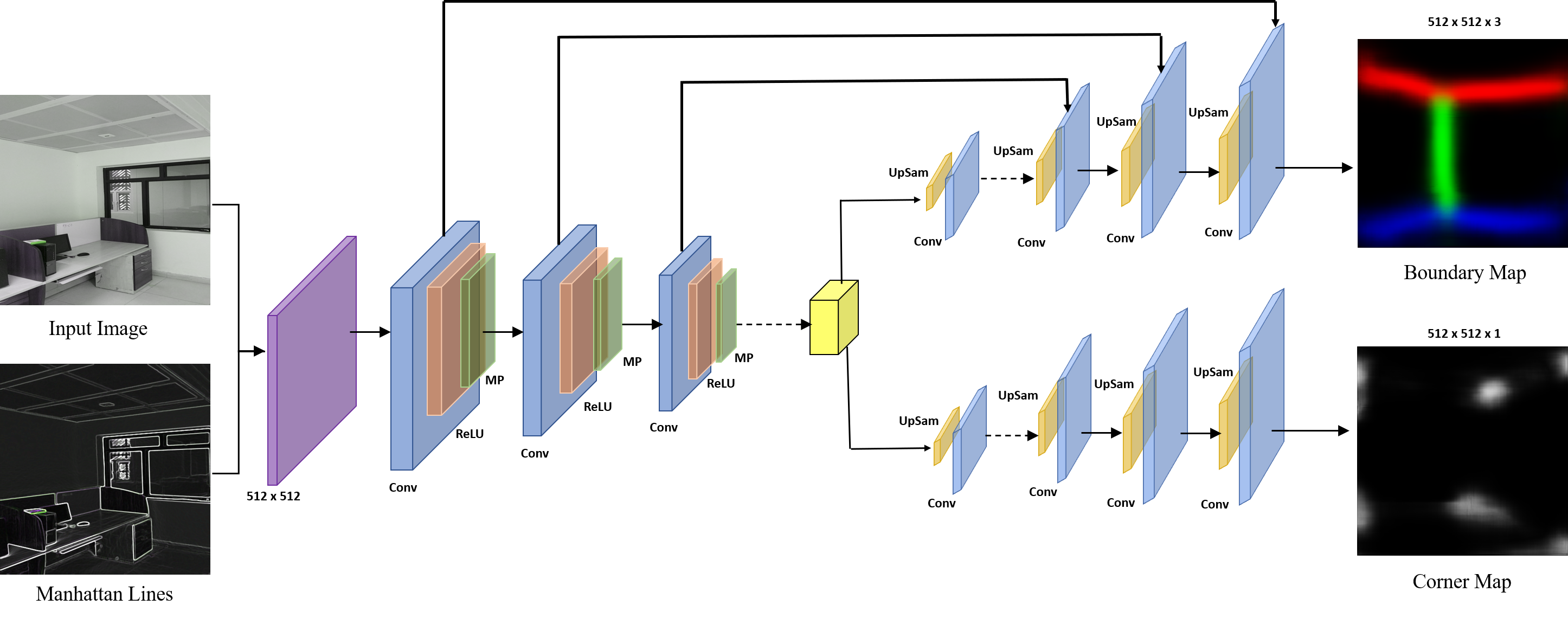}
    \caption{An illustration of how LayoutNet is adapted for the layout estimation in GRIHA.}
    \label{fig:layout}
\end{figure}

\subsection{Layout Estimation}
\label{sec:layout}
\begin{figure}[!b]
    \centering
    \includegraphics[width=\linewidth]{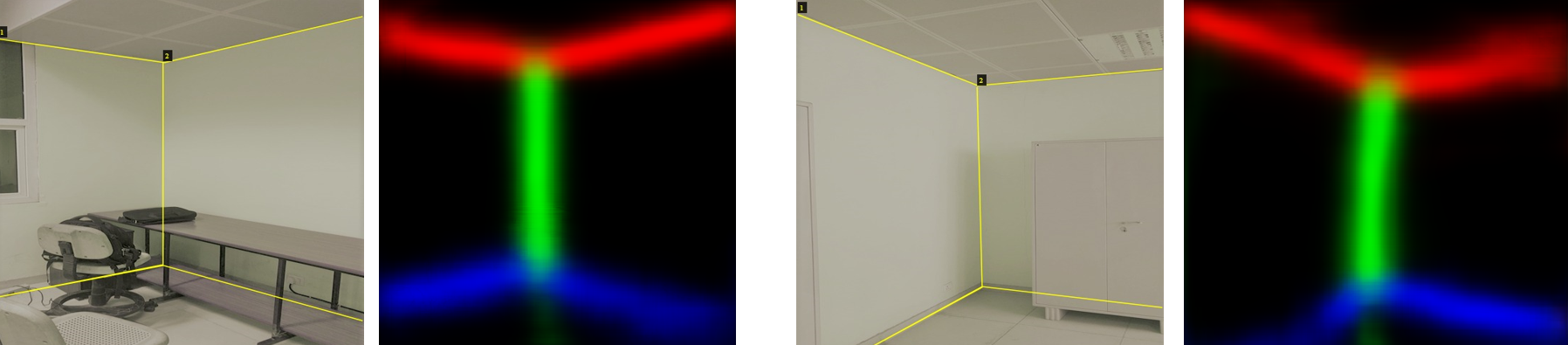}
    \caption{Results of layout generation on our dataset.}
    \label{fig:lay_res}
\end{figure}

While analyzing each partial scene, it is required to identify each pixel’s classification as a wall or edge pixel. This classification/segmentation helps in the identification of the layout of the indoor scene into consideration. Hence, edge and boundary maps of a scene is a requirement for layout estimation. In our work, we have adapted the technique proposed in \cite{zou2018layoutnet} to identify the scene’s edge/boundary maps. Figure. \ref{fig:layout} shows the network architecture of LayoutNet and its inputs and respective outputs. LayoutNet network is an encoder-decoder structure. The encoder consists of seven convolutional layers with a filter size of $3 \times 3$ and ReLU (Rectified Linear Unit) function and max-pooling layer follow each convolutional layer. Decoder structure contains two branches, one for predicting boundary edge maps and the other for corner map prediction. Both decoders have similar architecture, containing seven layers of nearest neighbor up-sampling operation, each followed by a convolution layer with a kernel size of $3 \times 3$ with the final layer being the Sigmoid layer. Corner map predictor decoder additionally has skip connections from the top branch for each convolution layer. Since their FOV (Field Of View) of the images is smaller, an additional predictor for predicting room type is added to improve corner prediction performance. The RGB image of the scene also takes Manhattan line segments as additional input that provides other input features and improves the network’s performance. Figure \ref{fig:lay_res} shows the predicted edge and corner maps for the input image on our dataset. All the annotations for performance evaluation is done using the annotation tool proposed in \cite{dutta2019vgg}. Our dataset’s images also include images that have occluded corners and wall edges, so there is no requirement for the manual addition of corners.

\subsection{3D Reconstruction}
\label{sec:3d}
To generate a layout of the entire scene, a 3D reconstruction of each partial scene must be done. This step is not required in the methods where depth cameras or specialized hardware are used to capture the images/point clouds. However, in the proposed work, RGB images are taken from a mobile phone's camera, which does not have the first place’s depth information. Hence, 3D reconstruction of each scene image is required to be done since we have tried to suppress additional hardware requirements for this task. We mapped every pixel of the RGB image with depth maps generated in the previous steps to create a point cloud for the scene. This step is preceded by a camera calibration step to identify the phone's camera’s intrinsic parameters: focal length $(f)$, center coordinates $(C_x, C_y)$. Here, coordinates in the point cloud are generated by:

\begin{align}
    Z & = \frac{D_{u,v}}{S}\\
    X & = \frac{(u- C_x)*Z}{f} \\
    Y & = \frac{(v- C_y)*Z}{f}
\end{align}

Here, $X,Y,Z$ are coordinates corresponding to the real world and $Z$ is the depth value. $D_{u,v}$ is the depth value corresponding to the $(u,v)$ pixel in the image in the depth map. $S$ is the scaling factor of the scene, obtained empirically and comparing dimensions of real-world objects and point clouds.Here, $f$, $C_x$, $C_y$ are the intrinsic parameters of the camera, generated by calibration. At the end of this step, we have 3D reconstruction of each partial scene in the real world scale ( Fig. \ref{fig:framework}(d)). The red marker shows the pose of the camera while capturing the RGB image in the scene.

\begin{figure*}[b]
    \centering
    \includegraphics[width=\linewidth]{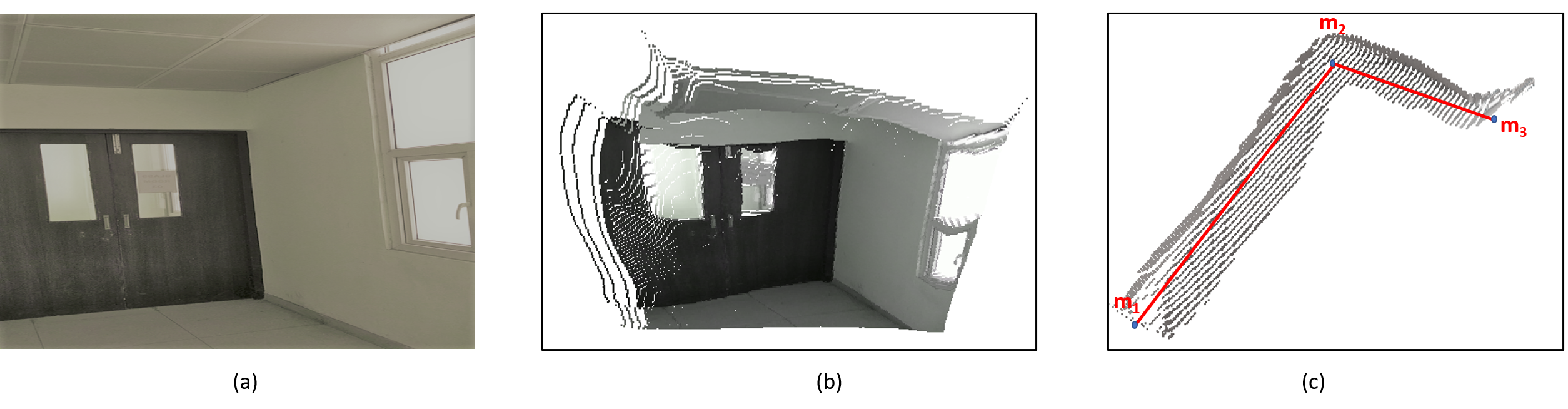}
    \caption{An illustration of partial scene layout generation from 3-D point cloud.}
    \label{fig:partial}
\end{figure*}
\subsection{Point cloud transformation and regularization}
The generated point clouds in the previous step are then mapped with the edge maps generated to identify the boundary pixels in the point cloud and projected in 2D space (as illustrated in Fig. \ref{fig:framework}(e)). These edge maps (as discussed in Sec. \ref{sec:layout}) are used to classify the pixels in point clouds in-wall and edge classes to identify the room’s geometry. These point clouds are scattered 3D points of the layout and required to be regularized to reduce the error in the generated 2D layout’s geometry (as illustrated in Fig. \ref{fig:partial} for a given room). We utilized ARCore to extract the 3D pose and generated the camera trajectory from the extracted pose, which is depicted by dotted arrows, as shown in the top panel of Fig. \ref{fig:framework}(e). It returns rotational and translation coordinates for each scene image for the first images taken. All the captured images are mapped to the local coordinate system of the image acquisition application discussed in Sec. \ref{sec:data_collection}. Also, there is no requirement of rotating the coordinate system while considering the transformation.

\begin{algorithm}[hbt!]
\caption{Regularize point clouds (PC)}\label{alg:reg}
\begin{algorithmic}[1]
\State {$\forall{R_j} \in \mathbf{R}$}  \Comment{$\mathbf{R}$: Total number of rooms}
\For{$i=1:n$}  \Comment{$n$: no of PC} 
      \State $P_i= 2D Point Clouds$  \label{ln:1a}
      \State $K= boundary(P_i)$  \label{ln:2}
      \State $C(c_1,c_2,...,c_k)= kmeans(P_i(K))$ \Comment{$C$: Clusters} \label{ln:31}
      \State $m_1, m_2, m_3 = mean(c_1), mean(c_2), mean(c_3)$  \label{ln:41}
      \State $line_1= line(m_1,m_2)$  \label{ln:51}
      \State $line_2= line(m_2,m_3)$  \label{ln:61}
      \While{$angle(line_1, line_2)<= 90$}  \label{ln:71}
      \State $Rotate(line_2)$  \label{ln:81}
      \EndWhile
      \State $RP_i= (line_1, line_2)$ \Comment{$RP_i$: local PC} \label{ln:91}
      \State $TP_i= (Rot({\theta_x,\theta_y,\theta_z})*Tr(t_x,t_y))* RP_i$   \label{ln:101}
    
\EndFor
\State $FP= polygon(TP_1, TP_2,...,TP_n)$ \Comment{$FP$: Final PC} \label{ln:111}
\For{$i=1:p$} \Comment{$p$: no of sides of polygon} 
     \State $\phi= angle(s_i,s_{i+1})$ \Comment{$s$: sides of polygon} \label{ln:121}
     \If{$\phi>90$ or  $\phi<90$}
     \State  $\phi(s_i,s_{i+1})= 0$ \label{ln:131}
     \EndIf
\EndFor
\end{algorithmic}
\end{algorithm}

Algorithm \ref{alg:reg} regularizes the local point cloud of each partial scene image for every room($R_j$) in all the room in a scene dataset ($\mathbf{R}$). Here, $P_i$ is the point cloud of each $i$-th scene where $n$ is the total number of point clouds. Boundary points for each $P_i$ is extracted in $P_i(K)$. Using the $k$-means algorithm, clusters of point set are made for $k=3$ on the Euclidean distance between them, where $m_1, m_2, m_3$ are the cluster means (line \ref{ln:41}). Since we are assuming the Manhattan world for the scene, the lines joining means are re-adjusted to have a right angle (line \ref{ln:81}). Each regularized local point cloud ($RP_i$) is transformed ($TP_i$) using rotation angle $\theta_x,\theta_y,\theta_z$, along each $x,y,z$ axis and translation coordinates $[t_x,t_y]$ returned by ARcore (line \ref{ln:101}). For global regularization, using each transformed point cloud, polygon ($FP$) is formed (line \ref{ln:111}), with $p$ number of sides ($s$). For each pair of sides, the angle between them ($\phi$) is checked and if they are not perpendicular, they are made collinear (line \ref{ln:131}) assuming the world to be Manhattan.

\begin{figure}[!b]
    \centering
    \includegraphics[width=\linewidth]{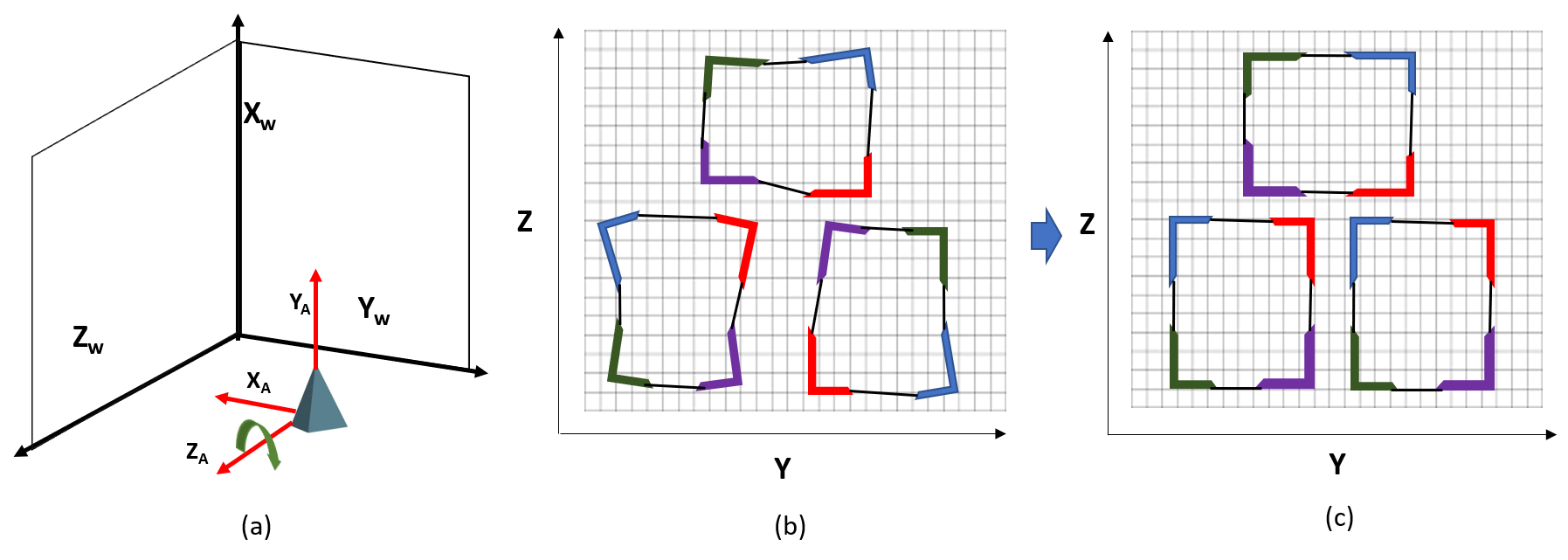}
    \caption{Depiction of translation and global regularization.}
    \label{fig:trans}
\end{figure}

 \begin{figure}[!b]
 \centering
 \subfigure[Finding the points which lie inside the boundary polygon.] {\includegraphics[width=0.45\linewidth]{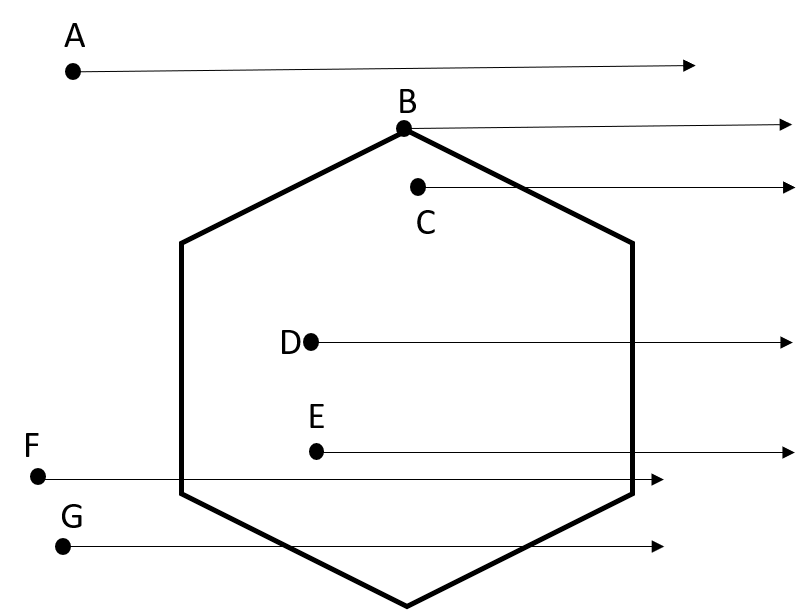}}
 \subfigure[Aligning the rooms with the boundary.] {\includegraphics[width=0.45\linewidth]{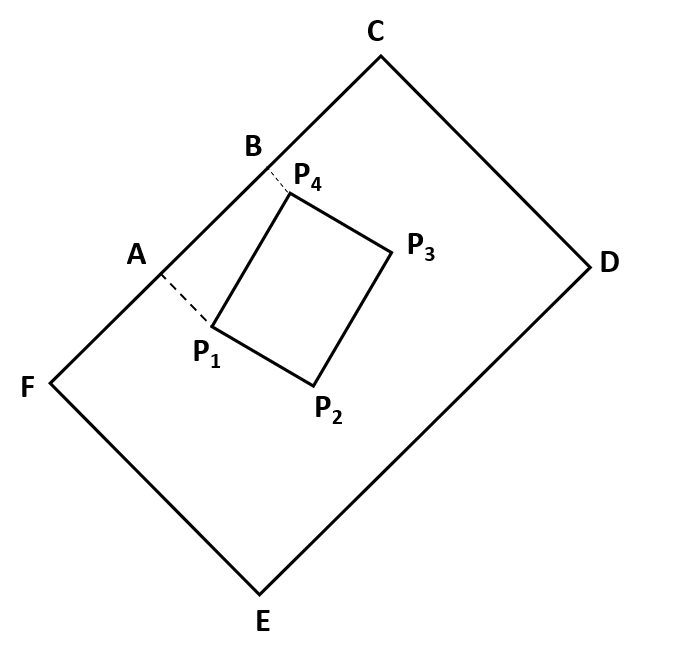}}
  \caption{An illustration of the intermediate stages of the regularization process.}
  \label{fig:geometry}
 \end{figure}

GRIHA performs regularization in $3$ steps. In step $1$, unique layouts are regularized, and in steps $2$ and $3$, the complete floor plan is regularized, taking everything collectively. Figure \ref{fig:partial} depicts the partial indoor scene with 2D layout generation. Figure \ref{fig:partial}(a) is the partial indoor scene RGB image into consideration, Fig. \ref{fig:partial}(b) is its 3D reconstruction in the form of point cloud, and Fig. \ref{fig:partial}(c) is the layout extracted from a 3D reconstructed point cloud using the previous steps’ layout. It shows the 2D projection of the layout point cloud, where $m_1$, $m_2$, $m_3$ are the means of three clusters extracted. It also shows the lines joining $m_1$, $m_2$, and $m_3$, the regularized point cloud from the projected set of points. 
 
 \begin{algorithm}[t]
 \caption{Finding the points inside the boundary polygon}
 \label{alg:reg2}
 
 \begin{algorithmic}[1]
 \For{$i=1:n$}
        \State $L_i= (P_i, \infty)$ \label{ln2:1}
         \If{$intersection(L_i, C_{hull}) == even$ } \Comment{$C_{hull}:$ Convex hull forming boundary} \label{ln2:2}
        \State $P_i \gets P_{outside}$  \Comment{$P_{outside}: $ Pool of points outside the boundary}
         \Else
        \State $P_i \gets P_{inside} $ \Comment{$P_{inside}: $ Pool of points inside the boundary}
        \EndIf
\EndFor    
 \end{algorithmic}
 \end{algorithm}

  \begin{algorithm}[!b]
 \caption{Aligning the points of each polygon to the boundary} 
 \label{alg:reg3}
 
 \begin{algorithmic}[1]
 \State $\forall P_i \in polygon  $
\State $L_1= P_i \perp CF$
\State $L_2= CF$
\State find equation of each line
\State $y= Y_C + m_{L_1}*(x-X_C) $
\State $m_{L_1}=  \frac{(Y_C- Y_F)}{(X_C-X_F)} $ \label{ln:7}
\State $m_{L_2}=  \frac{(Y_{P_i}- Y_A)}{(X_{P_i}-X_A)}$ \label{ln:8}
\State $m_{L_1}* m_{L_2}= -1$ \Comment{Perpendicularity condition} \label{ln:9}
\State $X_A= X_{P_i} + \frac{Y_C-Y_A}{X_C-X_F}* (Y_{P_i}-Y_A)$ \Comment{Substituting the known values to find unknowns } \label{ln:10}
\State $Y_A= Y_C + m_{L_1}*(X_A-X_C) $ \label{ln:11}
\State $X_{P_i}= X_A$ \Comment{Replacing the points of polygon with respective points on boundary} \label{ln:12}
\State $Y_{P_i}= Y_A$ \label{ln:13}

 \end{algorithmic}
 \end{algorithm}

\begin{figure*}[t]
    \centering
    \includegraphics[width=\linewidth]{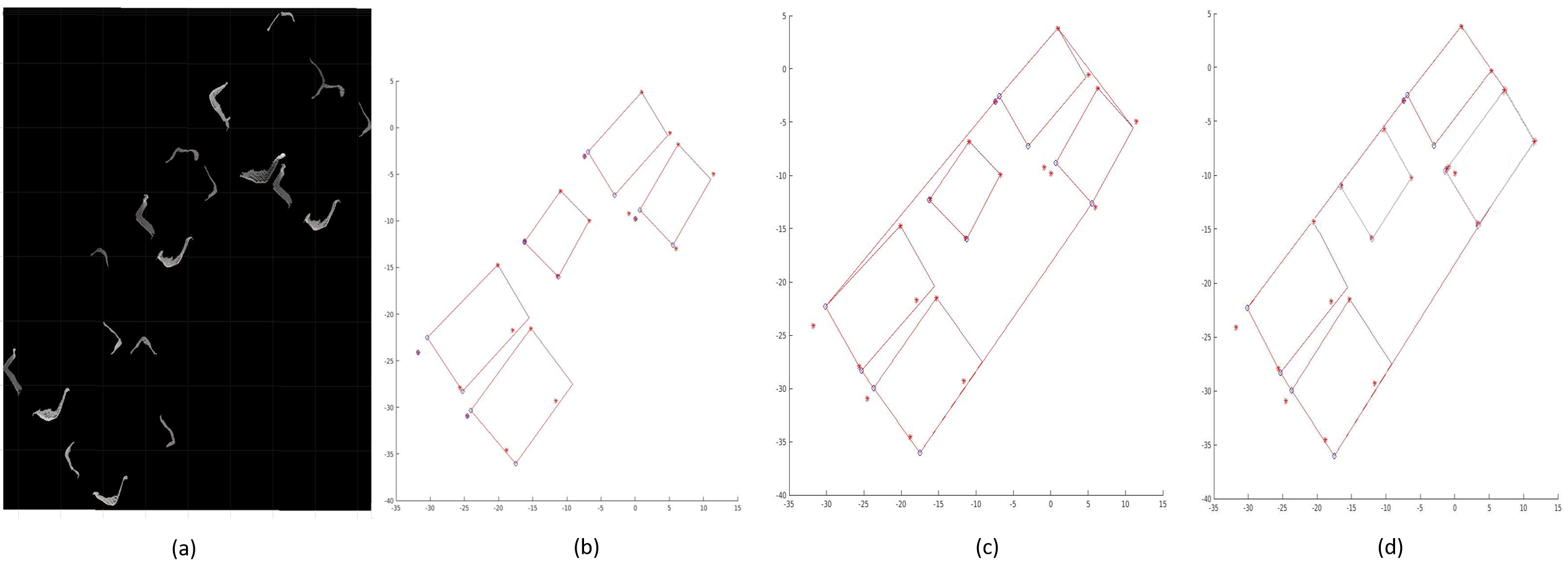}
    \caption{Step by step generation of the floor plan of an indoor scene.}
    \label{fig:process}
\end{figure*}

Figure \ref{fig:trans} depicts the transformation of partial point clouds and regularization of global layout. Figure. \ref{fig:trans} (a) shows the coordinate system in real-world ($X_W,Y_W,Z_W$) and in ARcore with the mobile device ($X_A,Y_A,Z_A$). ARcore transformations have to be rotated about the $Z_A$ axis to align the coordinate systems. Each partial 2D point set is then rotated and translated with the transformation given by ARcore (Fig. \ref{fig:trans} (b)). Figure \ref{fig:trans}(c) shows the globally regularized 2D point set for the partial point clouds for a set of rooms, and it agrees with the real world dimensions.

The further steps involved in the floor plan’s regularization include (1) generating boundaries for all the rooms and (2) a post-processing step, which aligns all the rooms with the boundary generated. Algorithm \ref{alg:reg2} and Alg. \ref{alg:reg3} depicts the process of finding a boundary for all the regularized layouts and their post-processing to align them along the boundary polygon, respectively. Algorithm \ref{alg:reg2} identifies the points for each room polygons inside the boundary polygon or on the boundary so that room polygons could be aligned with the boundary. Points that are supposed to be on the boundary but lying inside are identified using this algorithm. In Alg. \ref{alg:reg2}, line \ref{ln2:1}, a line $L_i$ is traced for each point $P_i$ to $\infty$, where line \ref{ln2:2} checks if the intersection of line $L_i$ with the boundary of Convex hull $C_{hull}$ is even number of times or odd number of times. If the intersection has happened $0$ or even number of times, then the point is considered to be outside the boundary, otherwise it is considered to be inside or on the boundary. 
Figure \ref{fig:geometry} (a) illustrates an example of such process. For each room polygon, points closer to the boundary line are identified, and a line is drawn from that point to infinity. If the line intersects the boundary polygon $0$ or even times (point A, F, G), then that point is outside the polygon. Otherwise, it is inside or on the polygon boundary (point B, C, D, E). The purpose of using Alg. \ref{alg:reg2} is to find the points which are lying inside the boundary and use them for further post-processing. If the point is identified to be inside the boundary polygon, then using Alg. \ref{alg:reg3}, they are aligned to the boundary line. 

Algorithm \ref{alg:reg3} shows the process of aligning the points of room polygons to the boundary polygon which are found to be inside. Figure. \ref{fig:geometry} (b) shows the example of the polygon $P_1 P_2 P_3 P_4$ which is required to be aligned with the boundary line $CF$. Points $P_1$ and $P_4$ are found to be inside the boundary polygon and needs to be aligned with line $CF$. Hence they need to be replaced with points $A$ and $B$ respectively.  Algorithm \ref{alg:reg3} finds the location of points $A$ and $B$ on line CF and replaces $P_1$ with $A$ and $P_4$ with $B$ by dropping a perpendicular line $P_1A$ on CF and using properties of perpendicular line segments for identifying the coordinates for $A$ and $B$. It checks slopes of both line segments (Alg. \ref{alg:reg3}, line \ref{ln:7} and line \ref{ln:8}) and checks the property of slopes of perpendicular line segments to identify $(X_A, Y_A)$ and $(X_B, Y_B)$, (Alg. \ref{alg:reg3}, line \ref{ln:9}). Once identified, it replaces both $P_1$ and $P_4$ with $A$ and $B$ (Alg. \ref{alg:reg3}, line \ref{ln:12} and line \ref{ln:13}). 

Figure \ref{fig:process} depicts the entire phases of generating a floor plan from its 2D projection to the final 2D floor plan. Figure \ref{fig:process}(a) depicts the 2D projection of each room’s partial point clouds on a floor after transformations taken by ARcore data. Figure \ref{fig:process}(b) shows the local and global regularized 2D layouts of each room, depicting the global relationship among them. In Fig. \ref{fig:process} (c), the boundary for all the rooms is generated by finding a convex hull for all the polygons and lines. Figure. \ref{fig:process}(d) shows the further refined and post-processed floor plan from the 2D projections. Further to the process of post-processing of the floor plans generated at this stage, door detection in the scenes and marking them in the floor plans is explained in the coming section.

\begin{figure}[t]
    \centering
    \includegraphics[width=\linewidth]{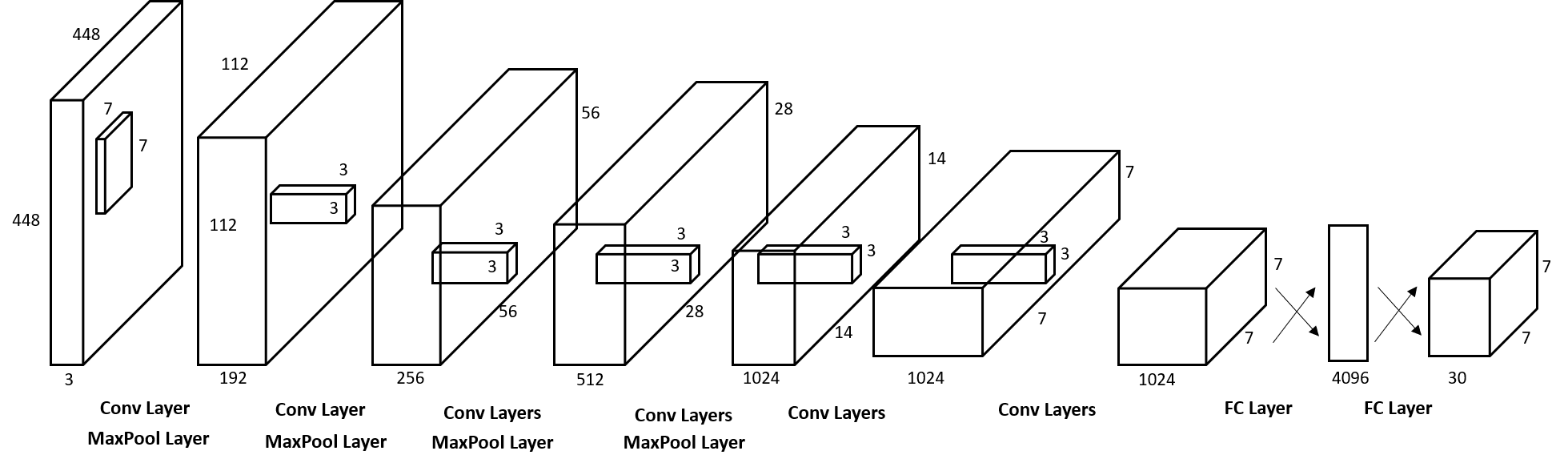}
    \caption{A schematic representation of the network architecture of YOLO for door detection.}
    \label{fig:door_YOLO}
\end{figure}

\subsection{Door Detection and placement}
\label{sec:door}
Indoor object detection such as doors, windows or other objects in indoor environment, from images or videos is a widely explored problem, solved using object detection networks such as YOLO, Faster-RCNN, SSD, etc. However, a dataset containing doors or windows specific to indoor scenes is not commonly available. In the current scenario, the post-processing has been limited to marking of doors, leaving others indoor objects due to very less variation in the other objects in the available dataset. It is challenging to generate a dataset containing doors in an indoor environment with diversity to train/ fine-tune existing networks. Hence, we used the DoorDetect dataset \cite{arduengo2019robust}. In our work, we have trained the YOLO object detection network on the DoorDetect dataset to detect doors in the indoor scenes to complete the floor plans. YOLO’s detection network has $24$ convolutional layers followed by $2$ fully connected layers (see Fig: \ref{fig:door_YOLO}). Each alternating convolutional layer has a reduction of feature space from its preceding layer. The network is pre-trained with ImageNet-$1000$ class dataset. 

 \begin{figure}[t]
 \centering
 \subfigure[Bounding box around the door.]{\includegraphics[width=0.3\linewidth]{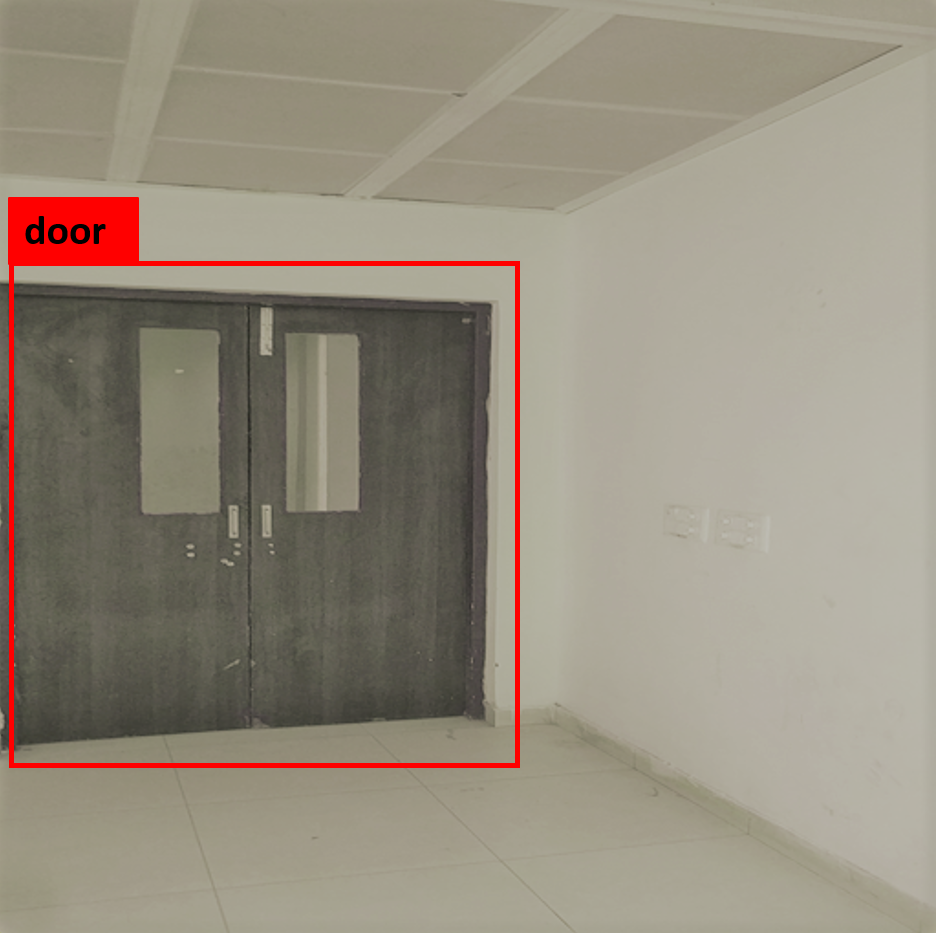}}
 \subfigure[Positioning of the doors on a floor plan.] {\includegraphics[width=0.3\linewidth]{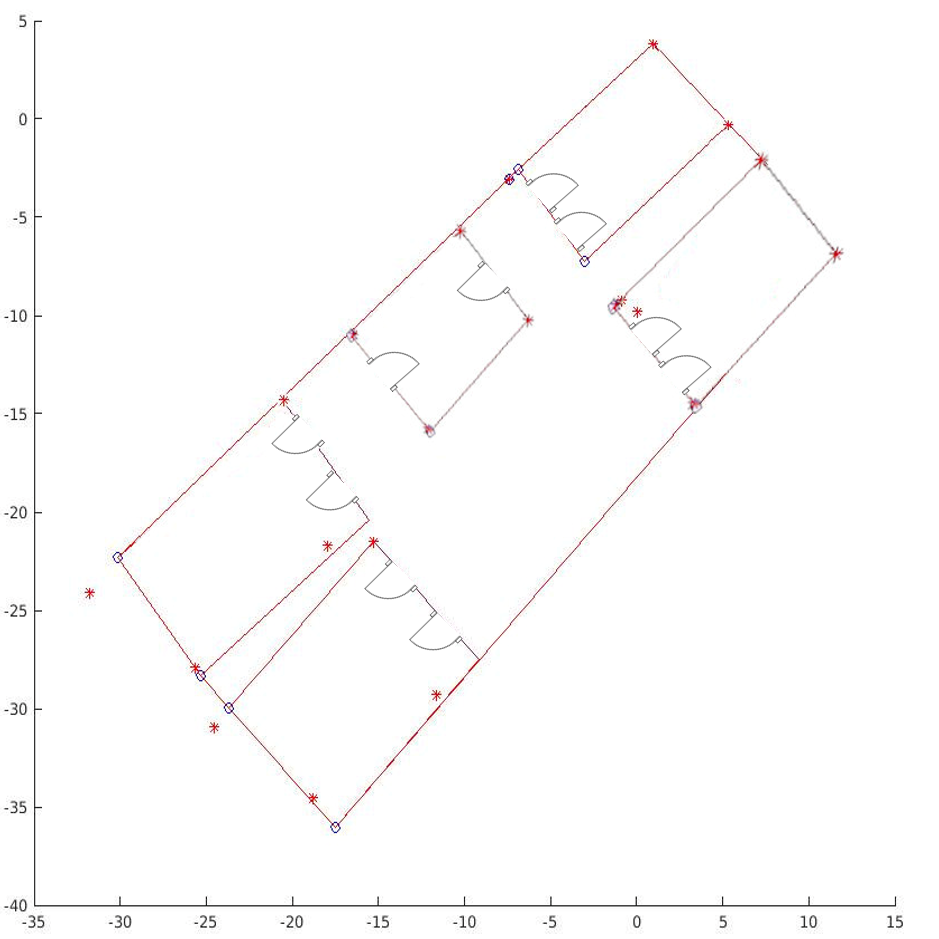}}
 \subfigure[Door centroid placement.] {\includegraphics[width=0.3\linewidth]{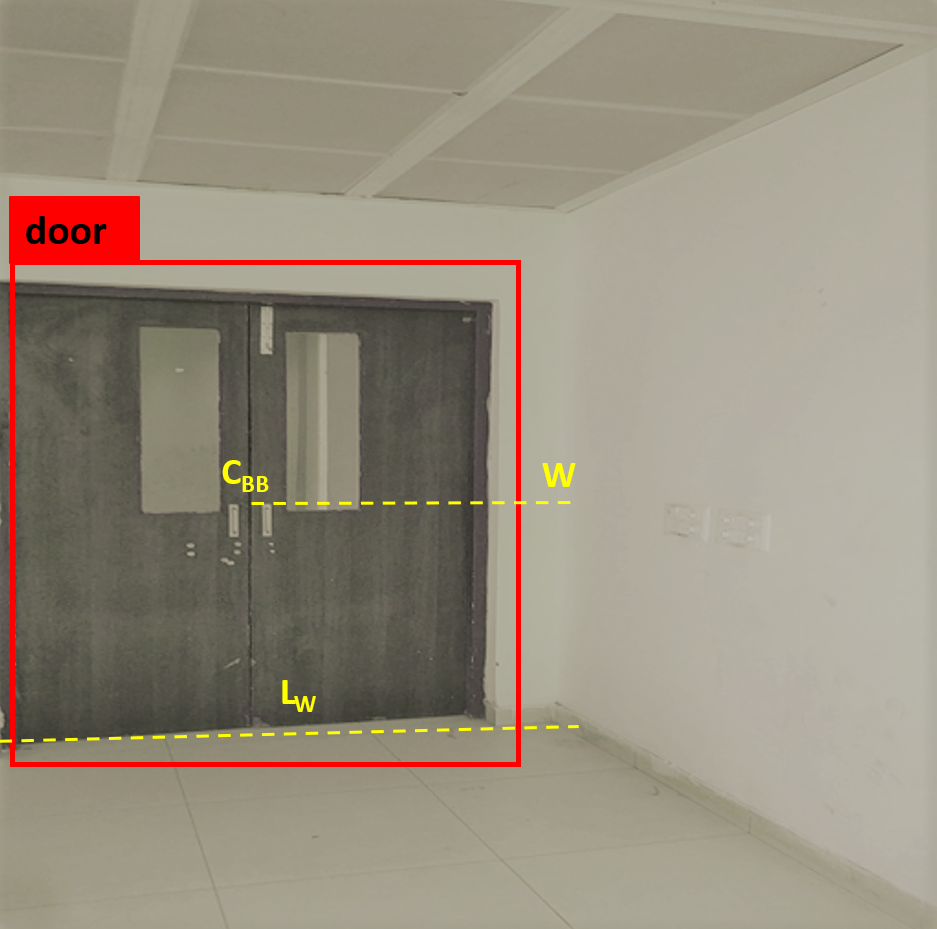}}
  \caption{An illustration of the performance of the door detection and placement algorithm. in floor plan.}
  \label{fig:door}
 \end{figure}
 
The DoorDetect dataset contains $1213$ images with annotated objects in an indoor environment. The door images contain various doors such as entrance doors, cabinet doors, refrigerator doors, etc. The mAP on DoorDetect dataset for YOLO came out to be $45\%$. Figure \ref{fig:door}(a) shows the door detected in one of the partial scene images in our experiments. Figure \ref{fig:door}(b) shows the door placement in the floor plan generated in the previous steps. Figure \ref{fig:door}(c) shows the parameters used for the door placement. The door placement is carried out using the following equations: 
\begin{eqnarray}
    Ratio_D=\frac{dist(C_{BB_I}, W_I)}{L_{W_I}}\\
     dist(C_{BB_F}, W_{I_F})= L_{W_F} * Ratio_D
\end{eqnarray}

Where $C_{BB_I}$ is the centroid of the bounding box of door detection (returned by door detection) in the real world image, $dist(C_{BB_I},W_I)$ is the distance between $C_{BB_I}$ and $W_I$ (wall), $L_{W_I}$ is the distance between two corners of the walls in the real world image  and $Ratio_D$ is the ratio between them. $Ratio_D$ is the ratio used for marking the doors in the generated floor plans with the reference of real world images of the scene. For each individual partial scene with a door, is marked with a respective door symbol in its corresponding floor plan. Here, $L_{W_F}$ is the distance between two corners of the walls in the corresponding 2D mapping (floor plan), $dist(C_{BB_F}, W_{I_F})$ is the distance between centroid of the door symbol  ($C_{BB_F}$) and wall ($W_{I_F}$) in the  corresponding 2D mapping (floor plan) which is the unknown entity and needs to be identified using $Ratio_D$ to mark the doors in the floor plan. 
 The axis of the door is kept perpendicular to the wall it belongs to. $Ratio_D$ is the ratio which is scale invariant for the generated floor plan and will remain same in the partial indoor scene image and its corresponding 2D map. In the next section, experimental findings will be discussed in details with their qualitative and quantitative comparison with state of the art methods. 

\section{Experimental Findings}
\label{sec:exp}
\subsection{Setup and Data Collection}

\begin{figure}[H]
    \centering
    \includegraphics[width=0.9\linewidth]{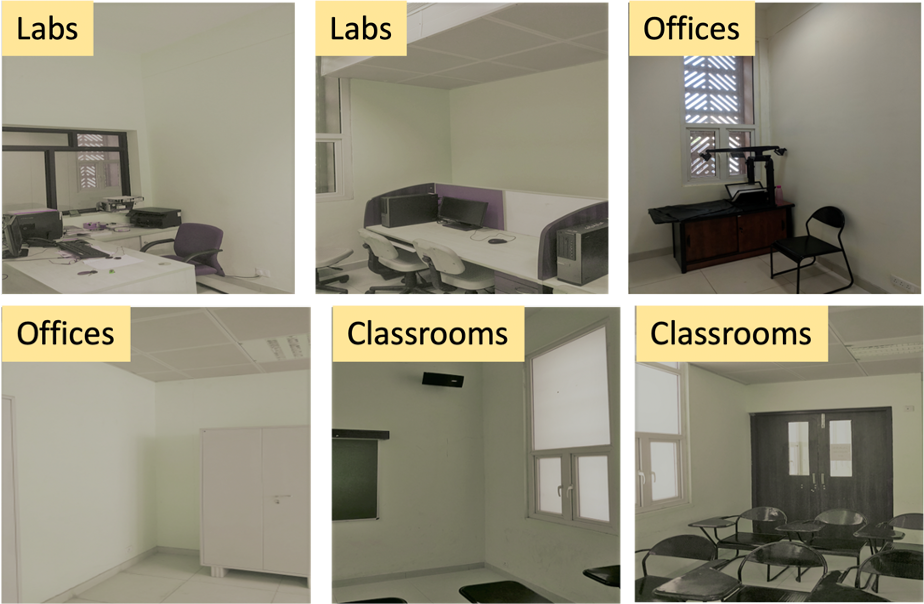}
    \caption{Sample images from the dataset.}
    \label{fig:dataset}
\end{figure}

For all the experiments, we have used two hardware platforms. They are Google Pixel $2$ XL and Samsung A50. We have utilized both of these mobile phones to deploy the data collection application and capture the images’ for all the locations. For depth estimation accuracy analysis on our dataset, structural similarity, and peak SNR metrics are used. Also, we have  used the metrics like pixel error and corner error for layout estimation accuracy analysis on our dataset. For evaluating the proposed layout estimation system’s performance, GRIHA, area, and aspect ratio error metrics are used as quantitative analysis. Qualitative analysis is also done to depict the proposed system’s robustness over existing Android and iOS based mobile applications. We have also compared the performance of GRIHA on the two hardware platforms mentioned at the beginning of this paragraph.  In GRIHA, we have performed experiments with a $3$ set of images. The first dataset is the right-wing of the ground floor of the Computer Science Department building in IIT Jodhpur, which are \textit{Classrooms}, the second is the left-wing of the same floor, which are \textit{Labs} and the third is the first floor of the same building which are \textit{Offices}. Figure \ref{fig:dataset} shows sample images from the collected data from each category. It can be seen that the images in the dataset can contain zero to moderate or heavy occlusion with differently illuminated environments.  

\begin{figure*}[t]
    \centering
    \includegraphics[width=0.9\linewidth]{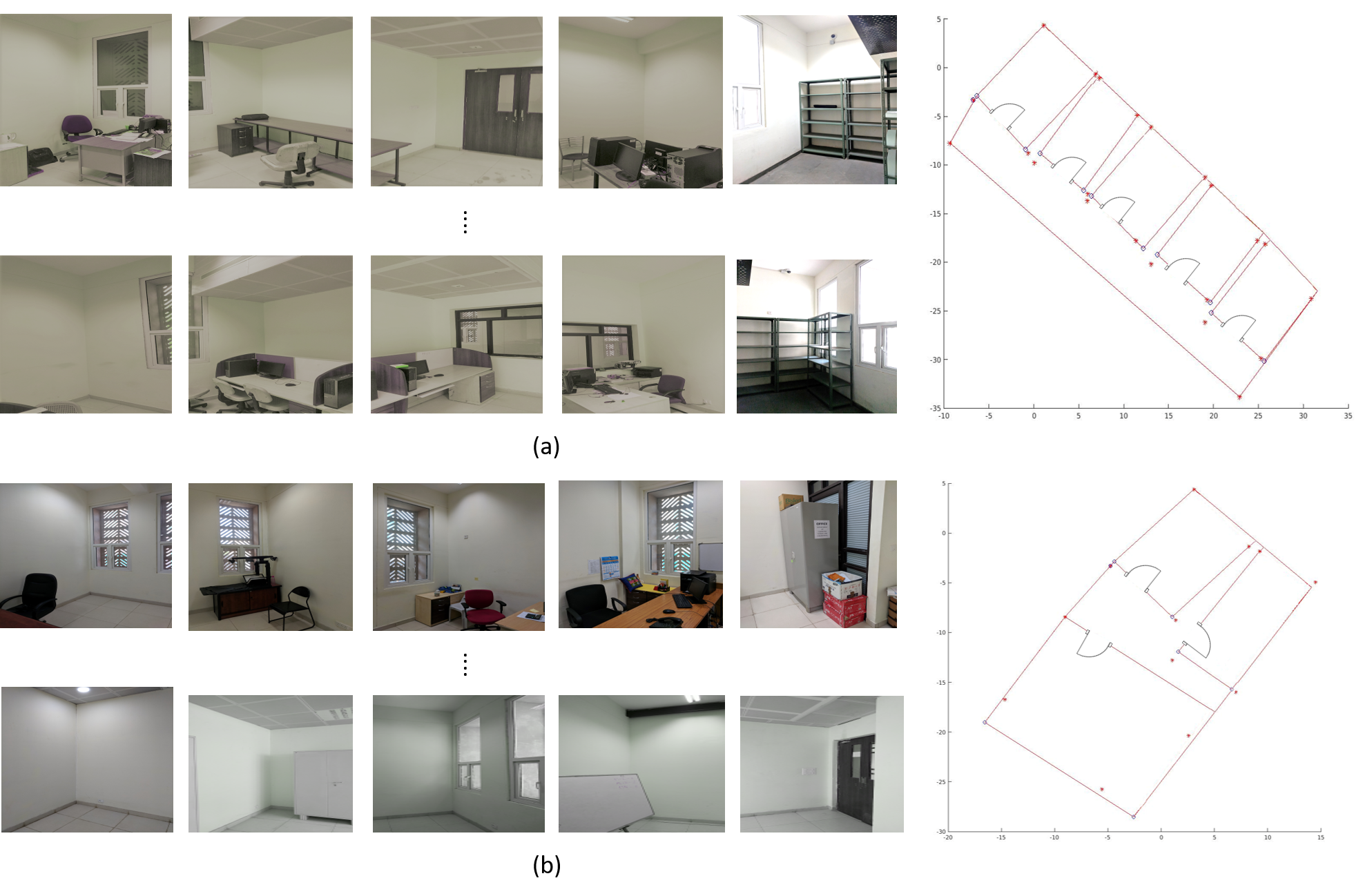}
    \caption{Estimated layouts for Labs and office datasets.}
    \label{fig:resultsoverall}
\end{figure*}
 
\subsection{Depth Estimations analysis}
Table \ref{tab:depth} shows the performance analysis of the depth estimation step in the proposed method. Ground truth depth maps for all the images in our dataset were generated using Kinect XBOX $360$ depth camera. The performance evaluation is done on two metrics,  Structural Similarity (SS) and peak SNR (PSNR) are defined as:  
\begin{eqnarray} \label{eq:SSI}
    SS(x,y)= \frac{(2\mu_x \mu_y + C_1 )* (2\sigma_{xy} + C_2)}{(\mu^2_x + \mu^2_y + C_2) * (\sigma^2_x + \sigma^2_y + C_2)}\\
    PSNR= 20 log_{10}* (MAX_I)- 10 log_{10}* (MSE)
\end{eqnarray}

In Eq. \ref{eq:SSI} , $\mu_x$ and $\mu_y$ are the mean intensity terms, while $\sigma_x$ and $\sigma_y$ are the standard deviations in the two image signals $x$ and $y$, $C_1$ \& $C_2$ are included to avoid instability when summations of mean intensities are close to zero. Also, for $PSNR$, $MSE$ is the mean square error  between the reference image and generated image, $MAX_I$ is the maximum possible pixel value of the image. Lower value of SS and PSNR indicates low quality of generated images as compared to reference ground truth image. It can be seen that the images in Labs dataset are performing worse than other dataset given its lowest value in terms of Structural Similarity, and PSNR because of the presence of variety of occlusion creating surfaces which creates irregular planes and limited field of view, making depth estimation a challenging task. As shown in Fig. \ref{fig:dataset}, for Labs scene images, the corners of partial scenes are highly occluded because of laboratory equipment, fixed sitting spaces and various other immovable heavy indoor objects.  

\begin{table}[t]
    \centering
    \caption{Performance analysis of depth estimation on our dataset}
    \begin{tabular}{||c|c|c|c||}
    \hline
        \textbf{Method} & \textbf{Classrooms} & \textbf{Labs} & \textbf{Offices} \\
        \hline
        Structural similarity &  $0.8433$  & $0.7528$&$0.8174$\\
        \hline
        Peak SNR&  $22.46$ & $17.55$ & $20.7808$\\
        \hline
         \hline
        
    \end{tabular}
    
    \label{tab:depth}
\end{table}

\begin{table}[!b]
    \centering
    \caption{Corner and edge estimation analysis on our dataset}
    \begin{tabular}{||c|c|c||}
    \hline
        \textbf{Scene} & \textbf{Corner error (\%) } & \textbf{Pixel error (\%)}  \\
        \hline
        Classrooms &  $1.04$  & $3.38$\\
        \hline
        Labs& $1.30$ & $4.15$\\
        \hline
        Offices & $1.12$ & $3.67$\\
         \hline
        
    \end{tabular}
    
    \label{tab:layout}
\end{table}

\subsection{Corner and edge estimation analysis}
Table \ref{tab:layout} shows GRIHA's performance on estimating the corners and edges of a room. The annotations for the layouts were generated using the tool proposed in \cite{dutta2019vgg}. The evaluation is done on two parameters, pixel error $\mathscr{P}$ and corner error. Pixel error identifies the classification accuracy of each pixel with the estimated layout and ground truth and averaged over all the images in a dataset. 
\begin{eqnarray}
    \mathscr{P}= \frac{\sum_{n=1}^{\mathbf{I_n}} Similarity(Pixel_E, Pixel_{GT})}{n}\\
    Similarity= 
\begin{cases}
    Pixel \in Edge Map,& \text{if } Pixel > 0\\
    Pixel \notin Edge Map,              & \text{Otherwise}
\end{cases}
    \label{eq:pixel}
\end{eqnarray}

where, $n$ is the total number of images in a dataset, $Pixel_E$ and $Pixel_{GT}$ are the pixels in estimated and ground truth images. Corner error $\mathscr{C}$ calculates the $L_2$ distance between the estimated corner and the ground truth corner of a room, normalized by image diagonal and averaged over all the images in a dataset. Here, $Corner_E$ and $Corner_{GT}$ are the estimated and ground truth corners. 

\begin{equation}
    \mathscr{C}= \frac{\sum_{n=1}^{\mathbf{I_n}} Dist_{L_2}(Corner_E, Corner_{GT})}{n}
    \label{eq:corner}
\end{equation}

It can be seen that Labs and Offices image dataset is more challenging than other datasets because of more occluded corners and edges with complex design of furniture and other experimental setups.

\subsection{Comparative Result}
Figure. \ref{fig:resultsoverall} shows the generated layouts for our collected image datasets. Figure. \ref{fig:resultsoverall} (a) shows resultant layout for labs dataset and Fig. \ref{fig:resultsoverall}(b) shows for offices dataset. Layout for labs dataset also include formation of corridor in the final layout, where the right panel shows the input stream of RGB images for the respective scenes. 

 \begin{table*}[t]
    \centering
    \caption{Quantitative evaluation of the estimated layouts for different scenes(S) and methods(M)}
    \resizebox{\textwidth}{!}{
    \begin{tabular}{|c|c|c|c|c|c|c|}
    \hline
       \textbf{Scene} $\longrightarrow$ & \multicolumn{2}{c|}{$\mathbf{Classrooms}$}  &\multicolumn{2}{c|}{ $\mathbf{Offices}$} &\multicolumn{2}{c|}{$\mathbf{Labs}$} \\
       \hline
       \textbf{Methods}$\downarrow$ &Area (E\%)&Aspect Ratio (E\%)&Area (E\%)&Aspect Ratio (E\%)&Area (E\%)&Aspect Ratio (E\%)\\
       \hline 
    GRIHA   & $\mathbf{3.12}$ & $\mathbf{2.21}$ &$\mathbf{3.25}$& $2.65$& $5.59$ &$3.07$ \\
    \hline
    Magic Plan \cite{magicplan} &${4.53}$&${3.34}$&${3.67}$&${1.81}$&${5.52}$&${3.03}$\\
    \hline
    Tape Measure \cite{tapemeasure} &${3.55}$&${3.58}$&${8.26}$&${1.71}$&${6.93}$&$\mathbf{1.21}$\\
    \hline
    Measure \cite{googleMeasure} &${7.27}$&${4.06}$&${6.65}$&${2.93}$&${6.02}$&${3.07}$\\
    \hline
    ARplan 3D \cite{arplan} &${3.15}$&${5.20}$&${4.40}$&$\mathbf{1.62}$&$\mathbf{4.39}$&${2.87}$\\
    \hline
    \end{tabular}}
    \label{tab:exp}
\end{table*}

A comparative study was performed with applications such as Magic Plan \cite{magicplan}, Tape Measure  \cite{tapemeasure}, Measure \cite{googleMeasure}, and AR Plan3D \cite{arplan} with the given ground truth measurements for each dataset. For every categories of images, the ground truth measurement is done by manually measuring each room’s dimension in each dataset and evaluating the area and aspect ratio, respectively. Quantitative evaluation is done on mean absolute $\%$ error for area and aspect ratio for each dataset. 

 \begin{equation}
     \textrm{Mean Absolute \% Error (E)}= \frac{1}{\mathbf{R}}\sum_{i=1}^{\mathbf{R}}\left|\frac{x_{GT}-x_i}{x_{GT}}\right|
     \label{eq: error}
 \end{equation}
 where \textbf{R} is the total number of rooms in a dataset, $x_i$ is the area/ aspect ratio of room $R_i$ and $x_{GT}$ is the ground truth area/ aspect ratio for the same.

 Table \ref{tab:exp} depicts the quantitative evaluation for the estimated layout for different scene dataset and other applications of Android and iOS. Results show that GRIHA performs best in terms of mean error $\%$ (E) in area and aspect ratio for Class Room dataset and area error for office dataset. For the lab dataset, ARplan3D performs best in terms of area error and Tape measure in aspect ratio error.

 \begin{table}[!b]
    \centering
    \caption{Qualitative comparison of GRIHA and state-of-the-art.}
    \begin{tabular}{|c|c|c|}
    \hline
        \textbf{Method} & \textbf{User Interaction} & \textbf{Manual Intervention} \\
        \hline
        GRIHA & $4$ Nos. & Not required\\
        \hline
        Magic Plan \cite{magicplan}& Continuous Scan & Add corners\\
        \hline
        Tape Measure \cite{tapemeasure} & Continuous Scan & Add corners\\
        \hline 
        Measure \cite{googleMeasure} & Continuous Scan & Add corners\\
        \hline
         ARplan 3D \cite{arplan} &  Continuous Scan & Add corners, height\\
         \hline
    \end{tabular}
    
    \label{tab:exp2}
\end{table}
 Table \ref{tab:exp2} depicts the qualitative comparison between the proposed method and other applications. Here, the number of user interactions and the amount of manual intervention required were considered based on the comparison. In terms of user interaction, our method requires only $4$ interactions, i.e., images of $4$ corners of a room, while other applications require a continuous scan and movement in the entire room. In terms of manual intervention, the proposed method does not require any after clicking the pictures. Whereas the other apps require manually adding the corners and height of the room. The proposed method’s only requirement is to click images, while other apps take some time and manual calibration to understand the environment and features. Due to this continuous scanning and more manual intervention, techniques like \cite{magicplan} yield more accurate results than ours. However, in the existing apps, if some object occludes the corner, the user has to add the corner themselves. A slight user error can heavily affect the accuracy of the layout. The accuracy of the existing apps also suffers in limited salient features in different scene frames while scanning.

 \begin{figure}[H]
 \centering
 \subfigure[ARplan 3D]{\includegraphics[width=0.3\linewidth]{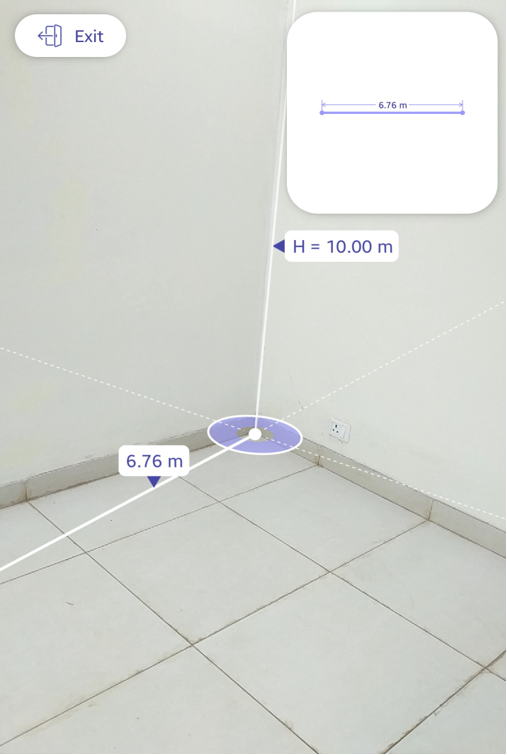}}
 \subfigure[Magic Plan] {\includegraphics[width=0.3\linewidth]{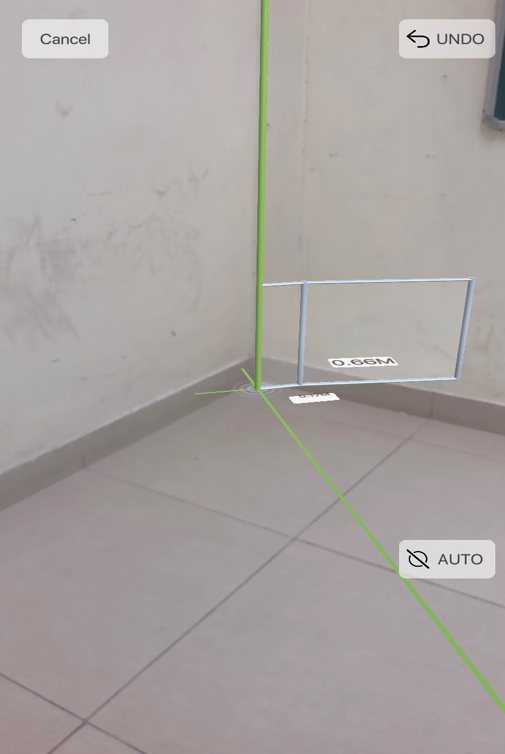}}
 \subfigure[Tape Measure] {\includegraphics[width=0.3\linewidth]{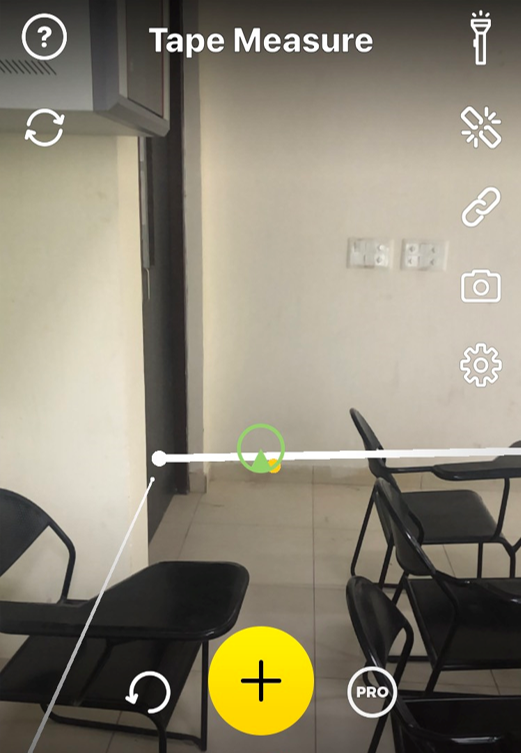}}
  \caption{Graphical User Interfaces (GUI) for different layout estimation applications.}
   \label{fig:robust}
 \end{figure}

\subsection{Robustness}
Figure. \ref{fig:robust} shows the screenshots of other publicly available mobile applications while collecting the dataset for evaluation. Figure. \ref{fig:robust}(a) is GUI for ARplan 3D,  Fig. \ref{fig:robust}(b) is for Magic Plan, and  Fig. \ref{fig:robust}(c) is for Tape Measure. There is a lot of manual interruption that requires layout estimation using these apps. For example, for ARplan 3D and Magic Plan, the rooms’ corners have to be added manually, which is a very error-prone process. The person holding the mobile device has to be cautious and accurate while adding the corners. Otherwise, there will be an error in the measurement of the edges of the room. Also, Fig. \ref{fig:robust}(c) shows that if the edges or corners of the room are occluded by furniture or other room fixtures, measuring the edges with these applications is impossible since the wall edges are invisible and have to be scanned through the furniture only.

However, in the proposed system GRIHA, these issues have been taken care of making it more robust than the existing mobile applications. GRIHA, do not require any manual interruption. Hence, the possibility of introducing any manual error is ruled out. Also, it does not require the mobile device to be run through all the room edges, making it easy for a user to use and robust in an occluded environment. The existing applications require some time after their launch and need a manual/automatic calibration of AR sensors by rotation and device scanning against the plane ground or wall. The automatic calibration by plane detection becomes difficult or takes longer when the room’s lighting condition is not proper or there is no difference in the color of the wall or ground. However, this is not a requirement in the proposed system. The user only requires to click images of the room, making it more robust in different lighting and interior environments. 
\begin{figure}[t]
    \centering
    \includegraphics[width= \linewidth]{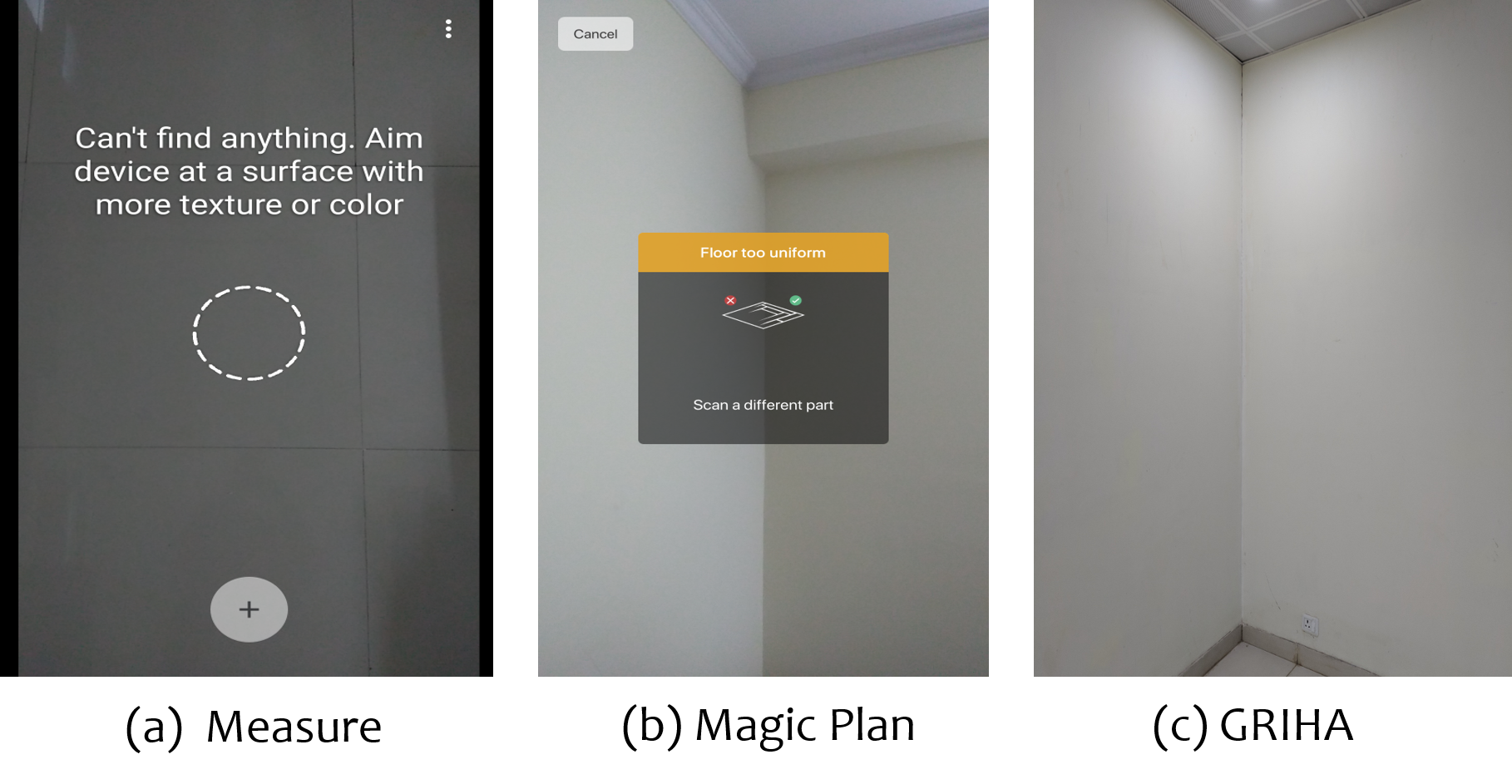}
    \caption{Scenes captured in different environments.}
    \label{fig:illumination}
\end{figure}

\begin{figure}[t]
    \begin{center}
\begin{tikzpicture}
\begin{groupplot}[
    group style={
    group name=my plots,
    group size=3 by 3, 
    xlabels at=edge bottom,
    ylabels at=edge left,
    horizontal sep=0.7cm, vertical sep=4cm,},
    legend style={at={(.5,0.9)},anchor=north east,legend columns=-1},
    ybar, /pgf/bar width=5pt,
    enlargelimits=0.05,
    ylabel={Error},
    symbolic x coords={GRIHA, Magic Plan, Tape Measure, ARplan 3D, Measure},
    xtick=data,
    x tick label style={rotate=45,anchor=east},
    nodes near coords,
    nodes near coords align={vertical}, 
    nodes near coords style={font=\tiny},
    width=0.4\linewidth,
]
\nextgroupplot[title=Classroom]
\addplot coordinates {(GRIHA,3.12) (Magic Plan,4.53) (Tape Measure, 3.55) (ARplan 3D, 3.15 ) (Measure, 7.27) };
\addplot coordinates {(GRIHA,3.4) (Magic Plan,4.12) (Tape Measure, 4.02) (ARplan 3D, 3.69) (Measure, 6.89) };
\nextgroupplot[title=Offices]
\addplot coordinates {(GRIHA,3.29) (Magic Plan,3.67) (Tape Measure, 8.26) (ARplan 3D, 4.4 ) (Measure, 6.65) };
\addplot coordinates {(GRIHA,3.59) (Magic Plan,4.2) (Tape Measure, 9.02) (ARplan 3D, 4.75) (Measure, 7.3) };
\nextgroupplot[title=Laboratory,legend to name=testLegend]
\addlegendentry{Google Pixel2 XL}
\addplot coordinates {(GRIHA,5.59) (Magic Plan,5.52) (Tape Measure, 6.93) (ARplan 3D, 4.39 ) (Measure, 6.02) };
\addlegendentry{Samsung Galaxy A50}
\addplot coordinates {(GRIHA,5.3) (Magic Plan,5.59) (Tape Measure, 7.3) (ARplan 3D, 4.83) (Measure, 6.63) };
\end{groupplot}
\end{tikzpicture}
\ref{testLegend}
\end{center}
    \caption{Comparative analysis of area error across devices.}
    \label{fig:area_error}
\end{figure}
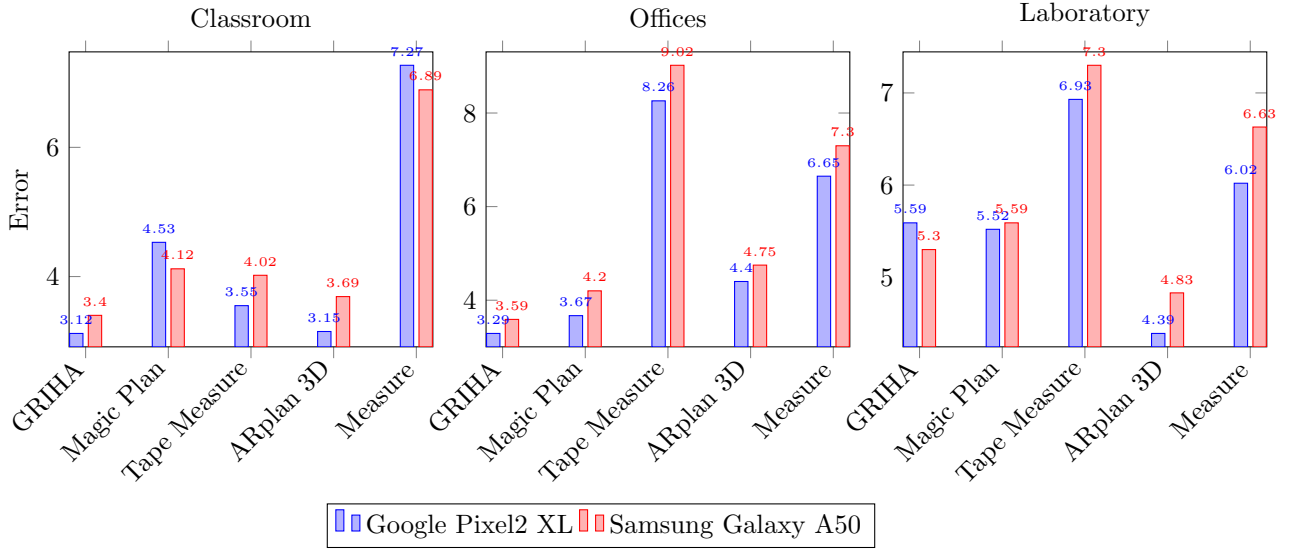
Different light conditions and environments affect the quality of images and final results of layout generation. In state of the art methods, different illuminated environment plays a key role in the functioning of the method. In poor illumination, different applications discussed in the previous section are not able to extract visual features. The applications require scanning of the entire scene with camera and require high contrast, edges and curved surfaces to detect feature points. If the captured images do not have enough feature points, then different keypoints and features are not detected. In poorly illuminated images there is a lack of contrast between two portions of a scene. Due to inconsistent indoor lighting they are not able to capture feature points and do not start functioning. In contrast, the proposed method GRIHA does not depend upon the illumination or high contrast surfaces in the captured images. Figure \ref{fig:illumination} depicts various scenes taken with mobile applications under different environments. Figure \ref{fig:illumination} (a) is taken with measure, under low illumination, in a smooth surface, which could not be scanned because keypoints could not be detected. Figure \ref{fig:illumination} (b) is taken with Magic Plan app in low light and consistent surface, which could not be scanned because of the same reason. However Fig. \ref{fig:illumination} (c) is a part of dataset collected in GRIHA, which is taken in low illumination and has a consistent surface of smooth walls, and it successfully contributed in the generation of layout. 

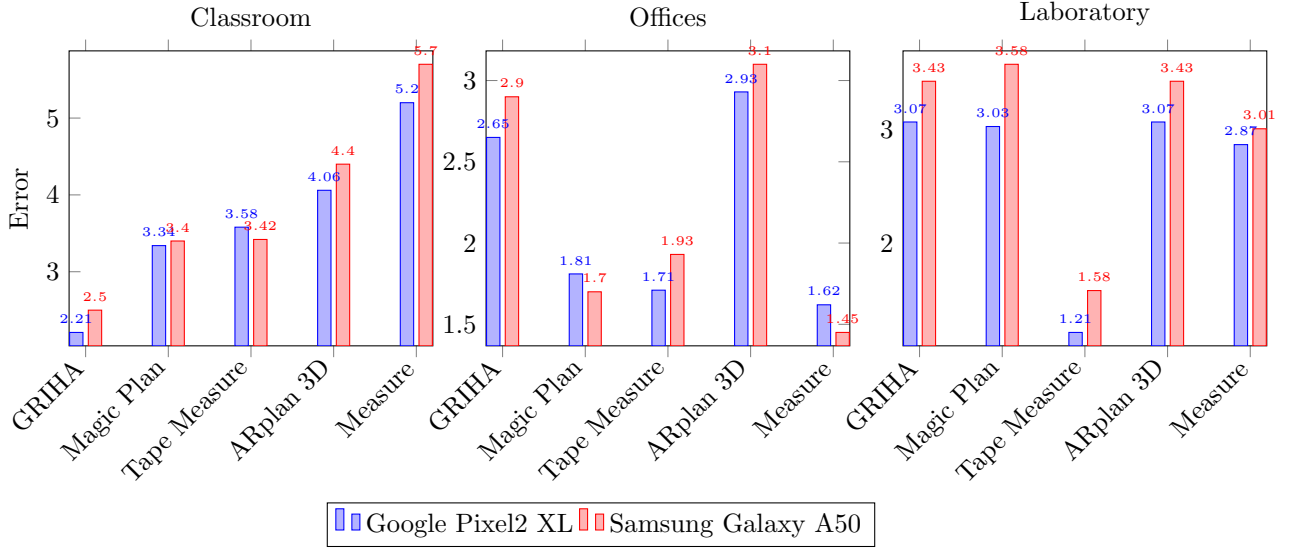
\begin{figure}[t]
    \begin{center}
\begin{tikzpicture}
\begin{groupplot}[
    group style={
    group name=my plots,
    group size=3 by 3, 
    xlabels at=edge bottom,
    ylabels at=edge left,
    horizontal sep=0.7cm, vertical sep=4cm,},
    legend style={at={(.5,0.9)},anchor=north east,legend columns=-1},
    ybar, /pgf/bar width=5pt,
    enlargelimits=0.05,
    ylabel={Error},
    symbolic x coords={GRIHA, Magic Plan, Tape Measure, ARplan 3D, Measure},
    xtick=data,
    x tick label style={rotate=45,anchor=east},
    nodes near coords,
    nodes near coords align={vertical}, 
    nodes near coords style={font=\tiny},
    width=0.4\linewidth,
]
\nextgroupplot[title=Classroom]
\addplot coordinates {(GRIHA,2.21) (Magic Plan,3.34) (Tape Measure, 3.58) (ARplan 3D, 4.06 ) (Measure, 5.2) };
\addplot coordinates {(GRIHA,2.5) (Magic Plan,3.4) (Tape Measure, 3.42) (ARplan 3D, 4.4) (Measure, 5.7) };
\nextgroupplot[title=Offices]
\addplot coordinates {(GRIHA,2.65) (Magic Plan,1.81) (Tape Measure, 1.71) (ARplan 3D, 2.93 ) (Measure, 1.62) };
\addplot coordinates {(GRIHA,2.9) (Magic Plan,1.7) (Tape Measure, 1.93) (ARplan 3D, 3.1) (Measure, 1.45) };
\nextgroupplot[title=Laboratory,legend to name=testLegend]
\addlegendentry{Google Pixel2 XL}
\addplot coordinates {(GRIHA,3.07) (Magic Plan,3.03) (Tape Measure, 1.21) (ARplan 3D, 3.07 ) (Measure, 2.87) };
\addlegendentry{Samsung Galaxy A50}
\addplot coordinates {(GRIHA,3.43) (Magic Plan,3.58) (Tape Measure, 1.58) (ARplan 3D, 3.43) (Measure, 3.01) };
\end{groupplot}
\end{tikzpicture}
\ref{testLegend}
\end{center}
    \caption{Comparison of aspect ratio error across devices.}
    \label{fig:aspect_error}
\end{figure}

\begin{figure}[!b]
    \centering
    \includegraphics[width=\linewidth]{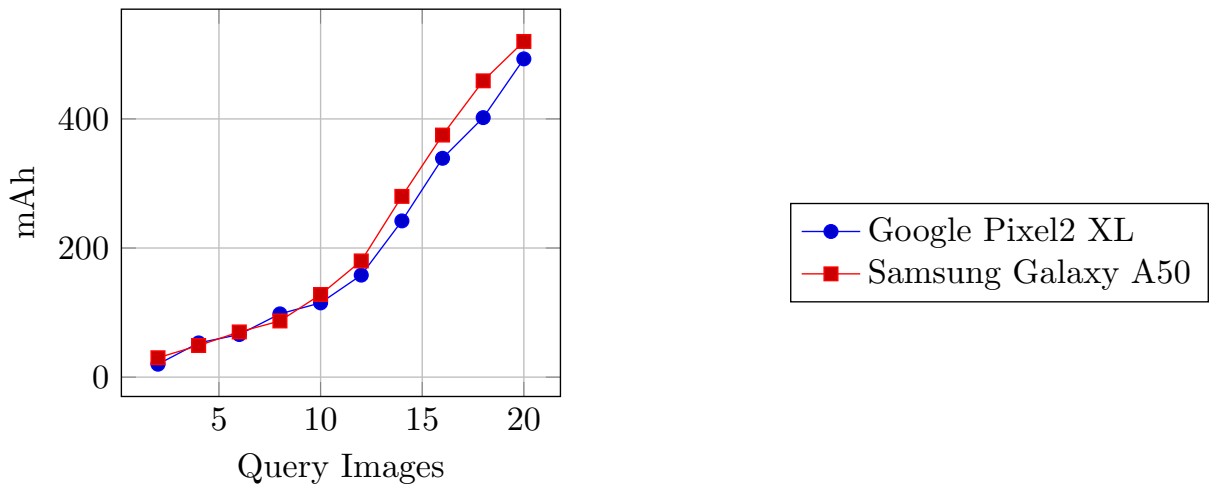}
    \caption{Comparison of power consumption across devices.}
    \label{fig:power}
\end{figure}

Figure. \ref{fig:area_error} and Fig. \ref{fig:aspect_error} shows the mean absolute error comparative analysis for area and aspect ratio across the two devices used. These plots infer the robustness and platform independence of the proposed system GRIHA. The relative performance of GRIHA is similar for both devices in terms of area error and aspect ratio error. 
Figure \ref{fig:power} shows the comparative analysis of power consumption in mAh with a growing number of query images across devices. It can be seen that Samsung Galaxy A50 is consuming more power for the proposed system and less efficient than Google Pixel 2 XL. Energy consumption is measured in terms of the battery used on each mobile device, which was recorded manually, from the start of data collection, with each query image collected.

State-of-the-art methods, such as \cite{zou2018layoutnet}, are also used to estimate the room’s entire floor plan. However, \cite{zou2018layoutnet} work on the panorama image of the scene. As compared to the proposed approach, \cite{zou2018layoutnet} requires more information to generate a layout. GRIHA works with non-overlapping pictures of the room, which do not have any shared features. Due to the lack of available datasets for partial scene images, we could not test our method on any publicly available datasets. We tested GRIHA on our dataset and performed experiments using the other apps under the same experimental condition to ensure fairness. GRIHA can generate a reasonably accurate layout in terms of the error in area, aspect ratio while requiring far less user interaction and intervention.

\section{Conclusion}
\label{sec:conclusion}
In this paper, we propose a technique, GRIHA, to generate room layout or floor plans using RGB images taken from the mobile phone's camera and camera pose data given by Google ARcore. Depth estimation and 3D reconstruction of the scene are done for the RGB images, and image stitching is done by transformation coordinates given by ARcore, assuming the world to be Manhattan. The estimated layout agrees well in terms of real-world dimensions. The proposed method works well for cluttered indoor settings and occluded walls and corners. The technique requires fewer user interactions and no manual intervention, as compared to other layout generation applications. We will try to estimate the entire floor’s layout by relaxing the Manhattan world assumption and for a more generalized scene in future work. 

 \section*{Acknowledgement}
 This work is partially supported by Science and Engineering Research Board, India under the project id ECR/2016/000953 and the Natural Sciences and Engineering Research Council of Canada (NSERC) under grant no. CRDPJ530666-18.

\bibliographystyle{spmpsci}


\begin{thebibliography}{31}
 \bibitem{arplan}{AR Plan 3D Ruler – Camera to Plan, Floorplanner.https://arplan-3d.en.aptoide.com/}
\bibitem{arcore}{ ARcore, fundamental concepts. https://developers.google.com/ar/discover/concepts}
 
\bibitem{googleMeasure}{Google Measure App. https://developer.android.com/}

\bibitem{tapemeasure}{ Occipital. The fastest way to measure (iOS application).https://tapmeasure.io/}
\bibitem{magicplan}{Sensopia.  MagicPlan. https://www.magicplan.app/magicplan/}
\bibitem{alhashim2018high}{ Alhashim,  I.,  Wonka,  P.:  High  quality  monocular  depth  estimation  via  transfer  learning.   arXivpreprint arXiv:1812.11941 (2018)}
\bibitem{angladon2018room}{Angladon,  V.:  Room  layout  estimation  on  mobile  devices.   Ph.D.  thesis,  University  of  Toulouse(2018)}
\bibitem{arduengo2019robust}{Arduengo, M., Torras, C., Sentis, L.: Robust and adaptive door operation with a mobile manipulator robot (2019)}
\bibitem{bao2014understanding}{ Bao, S.Y., Furlan, A., Fei-Fei, L., Savarese, S.:  Understanding the 3d layout of a cluttered roomfrom multiple images.  In:  WACV, pp. 690–697. IEEE (2014)}
\bibitem{cabral2014piecewise}{ Cabral, R., Furukawa, Y.:  Piecewise planar and compact floorplan reconstruction from images.  In:CVPR, pp. 628–635. IEEE (2014)}
\bibitem{chelani2018towards}{ Chelani, K., Sidhartha, C., Govindu, V.M.:  Towards automated floorplan generation (2018)}
\bibitem{chen2015crowd}{Chen,  S.,  Li,  M.,  Ren,  K.,  Qiao,  C.:  Crowd  map:  Accurate  reconstruction  of  indoor  floor  plansfrom crowdsourced sensor-rich videos.  In:  2015 IEEE 35th International conference on distributed computing systems, pp. 1–10. IEEE (2015)}
\bibitem{dasgupta2016delay}{ Dasgupta, S., Fang, K., Chen, K., Savarese, S.: Delay: Robust spatial layout estimation for cluttered indoor scenes.  In:  CVPR, pp. 616–624 (2016)}
\bibitem{dong2015imoon}{Dong, Jiang and Xiao, Yu and Noreikis, Marius and Ou, Zhonghong and Y J, Antti,: Using smartphones for image-based indoor navigation.  In:  Proceedings of the 13th ACM Conference on Embedded Networked Sensor Systems, pp. 85–97 (2015)}
\bibitem{dutta2019vgg}{Dutta, A., Zisserman, A.: The VIA annotation software for images, audio and video. In: Proceedings of the 27th ACM International Conference on Multimedia,  MM ’19. ACM, New York,  NY, USA(2019).  DOI 10.1145/3343031.3350535.  URLhttps://doi.org/10.1145/3343031.3350535}
\bibitem{fernandez2018layouts}{ Fernandez-Labrador, C., Perez-Yus, A., Lopez-Nicolas, G., Guerrero, J.J.:  Layouts from panoramic images with geometry and deep learning.  IEEE Robotics and Automation Letters3(4), 3153–3160(2018)}
\bibitem{furlan2013free}{ Furlan, A., Miller, S.D., Sorrenti, D.G., Li, F.F., Savarese, S.:  Free your camera:  3d indoor scene understanding from arbitrary camera motion.  In:  BMVC (2013)}
\bibitem{hsiao2019flat2layout}{ Hsiao, C.W., Sun, C., Sun, M., Chen, H.T.:  Flat2layout:  Flat representation for estimating layoutof general room types.  arXiv preprint arXiv:1905.12571 (2019)}
\bibitem{lin2018floorplan}{ Lin, C., Li, C., Furukawa, Y., Wang, W.:  Floorplan priors for joint camera pose and room layout estimation.  arXiv preprint arXiv:1812.06677 (2018)}
\bibitem{liu2015rent3d}{ Liu, C., Schwing, A.G., Kundu, K., Urtasun, R., Fidler, S.: Rent3d: Floor-plan priors for monocular layout estimation.  In:  CVPR, pp. 3413–3421 (2015)}
\bibitem{murali2017indoor}{Murali, S., Speciale, P., Oswald, M.R., Pollefeys, M.: Indoor scan2bim: Building information modelsof house interiors. In:  2017 IEEE/RSJ International Conference on Intelligent Robots and Systems(IROS), pp. 6126–6133. IEEE (2017)}
\bibitem{murez2020atlas}{ Murez, Z., van As, T., Bartolozzi, J., Sinha, A., Badrinarayanan, V., Rabinovich, A.:  Atlas:  End-to-end 3d scene reconstruction from posed images.  arXiv preprint arXiv:2003.10432 (2020)}
\bibitem{okorn2010toward}{Okorn,  B.,  Xiong,  X.,  Akinci,  B.,  Huber,  D.:   Toward  automated  modeling  of  floor  plans.   In:Proceedings of the symposium on 3D data processing, visualization and transmission, vol. 2 (2010)}
\bibitem{phalak2020scan2plan}{  Phalak, A., Badrinarayanan, V., Rabinovich, A.:  Scan2plan:  Efficient floorplan generation from 3dscans of indoor scenes.  arXiv preprint arXiv:2003.07356 (2020)}
\bibitem{sun2019horizonnet}{Sun, C., Hsiao, C.W., Sun, M., Chen, H.T.:  Horizonnet:  Learning room layout with 1d representa-tion and pano stretch data augmentation.  In:  CVPR, pp. 1047–1056 (2019}
\bibitem{turner2014floor}{ Turner,  E.,  Zakhor,  A.:   Floor  plan  generation  and  room  labeling  of  indoor  environments  fromlaser range data.  In:  2014 international conference on computer graphics theory and applications(GRAPP), pp. 1–12. IEEE (2014)}
\bibitem{xu2017pano2cad}{ Xu, J., Stenger, B., Kerola, T., Tung, T.:  Pano2cad:  Room layout from a single panorama image.In:  WACV, pp. 354–362. IEEE (2017)}
\bibitem{zhang2013estimating}{ Zhang, J., Kan, C., Schwing, A.G., Urtasun, R.:  Estimating the 3d layout of indoor scenes and its clutter from depth sensors.  In:  ICCV, pp. 1273–1280 (2013)}
\bibitem{zhang2019edge}{ Zhang,  W.,  Zhang,  W.,  Gu,  J.:  Edge-semantic  learning  strategy  for  layout  estimation  in  indoorenvironment.  IEEE-T on cybernetics (2019)}
\bibitem{zhang2014panocontext}{ Zhang, Y., Song, S., Tan, P., Xiao, J.:  Panocontext:  A whole-room 3d context model for panoramic scene understanding.  In:  ECCV, pp. 668–686. Springer (2014)}
\bibitem{zou2018layoutnet}{ Zou, C., Colburn, A., Shan, Q., Hoiem, D.:  Layoutnet:  Reconstructing the 3d room layout from asingle rgb image.  In:  CVPR, pp. 2051–2059 (2018)}
\end{thebibliography}
 
\end{document}